\documentclass{mbd2011}
\pdfoutput=1
\newif\ifpdf\ifx\pdfoutput\undefined\pdffalse\else\pdfoutput=1\pdftrue\fi

\title{Adaptive Locomotion of Multibody Snake-like Robot}

\author{Eugen Meister, Sergej Stepanenko, and Serge Kernbach}

\heading{Eugen Meister, Sergej Stepanenko and Serge Kernbach}

\address{Institute of Parallel and Distributed Systems\\
University of Stuttgart, Universit{\"a}tsstr.~38, D-70569 Stuttgart, Germany\\
e-mails: \url{Eugen.Meister@ipvs.uni-stuttgart.de}, \\
 \url{sergstepanenko@gmail.com},\\
\url{Serge.Kernbach@ipvs.uni-stuttgart.de}}


\abstract{This paper represents an adaptive rhythmic control for a snake-like robot with 25 degrees of freedom. The adaptive gait control is implemented in algorithmic way in simulation and on a real robot. We investigated behavioral and energetic properties of this control and a dynamics of different body segments. It turned out that despite using homogeneous generators, physical constraints have an inhomogeneous impact on neighbor body segments. By analytical modeling of such dynamics, it may result in heterogeneous coupling of oscillators for a rhythmic control and impact scalability and synchronization effects of gait pattern generators.}

\begin{document}

\section{INTRODUCTION}

Reconfigurable robotics recently attracted essential attention of researchers due to flexibility, developmental plasticity, behavioral and functional adaptability~\cite{Levi10}. Modular robots with a high number of degrees of freedom are advantageous for industrial~\cite{Kornienko1994}, underwater and space applications, however are challenging for control of macroscopic locomotion~\cite{Kernbach08}. There are two distinctive approaches for locomotive control in such multi-body systems: based on pre-destined trajectories, for example screw theory~\cite{Murray1994}; or based on rhythmic generators~\cite{sChevallier07}. In this paper we assume that independent reconfigurable modules form a multi-body snake-like organism, see Fig.~\ref{fig:SnakeHardware}, and we limit ourselves to adaptive locomotion control using rhythmic gait generator. The concept of rhythmic control can be represented as a periodic generator $\dot {\underline x} = \underline f (\underline x, \underline \alpha)$ and the coupling term $\phi(\underline x, \underline \beta)$ as e.g. a two ways coupled dynamic system with open boundaries of size $n$
\begin{eqnarray}
\label{eq:rhythmic}
 \dot {\underline x_1} &=& f (x_1, \underline \alpha) + \phi(x_2, \underline \beta), \nonumber \\ \nonumber
 \dot {\underline x_2} &=& f (x_2, \underline \alpha) + \phi(x_1, x_3, \underline \beta), \\
 \dot {\underline x_3} &=& f (x_3, \underline \alpha) + \phi(x_2, x_4, \underline \beta), \\ \nonumber
  ... &=& ... ,\\\nonumber
 \dot {\underline x_n} &=& f (x_n, \underline \alpha) + \phi(x_{n-1}, \underline \beta),
 \end{eqnarray}
where $\underline x$ are state variables, $\underline \alpha, \underline \beta$ are sets of control parameters and $i$ is associated with i-body segment. Goal of the rhythmic control can be thought of as adjusting the local coupling terms $\phi(\underline x, \underline \beta)$~\cite{Levi99} so that the system (\ref{eq:rhythmic}) demonstrate a global synchronous gait pattern. The most of the state of the art works on the  CML/CPG-based (Coupled Map Lattices/Central Pattern Generators) rhythmic control, e.g. \cite{Kaneko93}, consider the coupling as homogeneous in relation to neighbor body segments.

Implementing this approach in real multi-body systems, we face the problem of constraints, such as friction, limited motor torque, mass distribution over the whole system or a variation of mechanical couplings between segments. It results in two important consequences. First of all, due to complexity of analytic representation of such constraints, the focus of system's modeling is shifted to algorithmic approaches~\cite{c21}. Secondly, the same physical constraints have different impact on all segments of a multi-body system. The appeared different dynamics of segments can be associated with coupling terms and it leads to heterogeneous couplings between individual gait generators. This heterogeneity has an implicit form and appears dynamically during interactions between segments of a multi-body system and environment (in form of embodied cognition~\cite{c1}).

This work intends to demonstrate two above mentioned points: algorithmic approach for adaptive rhythmic gait generator and appeared inhomogeneity of coupling terms. This inhomogeneity can be introduced as heterogeneous coupling terms for analytic CML/CPG-based approaches and in turn create some open questions of scalability and synchronization effects in rhythmic gait generators for multi-body systems. The scenario of this work originates from the large-scale European projects ``SYMBRION'' \cite{symbrion} and ``REPLICATOR'' \cite{replicator} and is related to an optimal coverage of unknown area by a snake-like behavior with optical collision avoidance. This approach also addresses the problem of locomotive adaptability to the structure of environment and an energetic efficiency of control.

The paper is organized in the following way. In Section 2 we introduce the adaptive algorithm for directional collision avoidance for snake-like robots. The implementation in Matlab and results of the algorithm are demonstrated in Section 3. The algorithm was tested on a real snake robot shown in Section 4. The 3D locomotion gait analyses for snakes is explained in Section 5. The simulation of dynamics in Matlab/Simulink is demonstrated in the following Section 6. Finally, Section 7 concludes the work.

\section{ADAPTIVE ALGORITHM FOR RHYTHMIC GAIT GENERATION}
\label{sec:AdaptiveAlg}

\begin{figure}[h!]
\centering
\subfigure{\includegraphics[width=.7\textwidth]{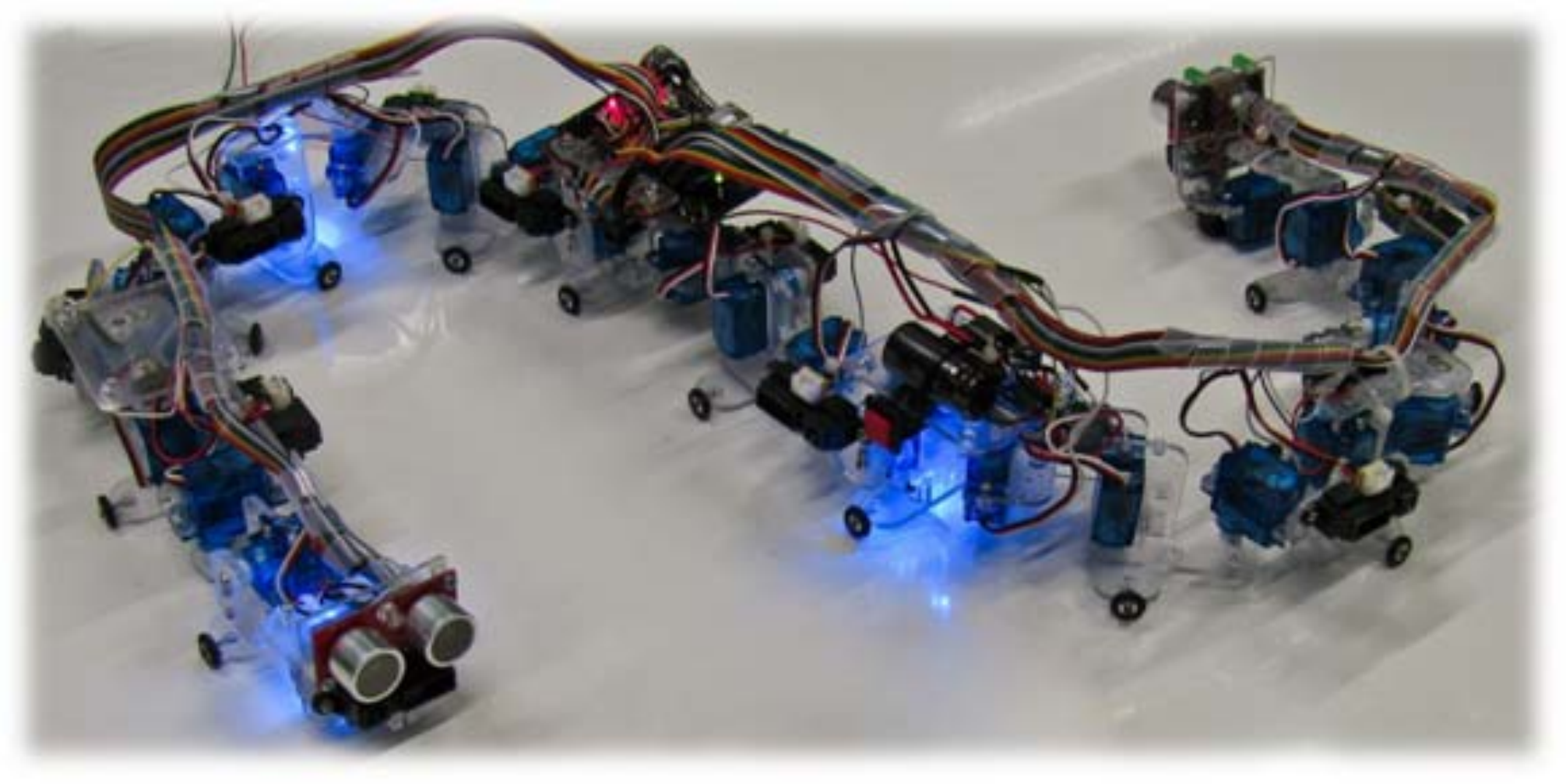}}
\caption{\it Snake-like robot platform with 25 DoF, 15 IR distance sensors, one front ultrasonic sensor, 3D accelerometer and two-axes compass.\label{fig:SnakeHardware}}
\end{figure}

For the snake-like robot, see Fig.~\ref{fig:SnakeHardware}, we use the serpentine gait often also called lateral undulation:
\begin{equation}
\varphi_{i,h}(t) = A_{h} sin(\omega_{h} t + (i-1)\phi_{h}) + \Upsilon_{h}, ~~~(i=1,...,n-1),
\label{eq:gate}
\end{equation}
where parameters $A_h$, $\phi_h$ determine the shape of the snake and the parameter $\Upsilon_h$ modifies the serpentine curve. Links are labeled by index $i$ and $h$ means that the gait acts in horizontal direction. The expression (\ref{eq:gate}) represents the state-of-the-art solution~\cite{Saito}, well known in robotics \cite{Hirose}, \cite{Burdick}, \cite{Ostrowski} and is inspired by efficient locomotive behavior of real snakes. Since the embedded platform provides a limited computational capability, our goal was to reduce the complexity of control mechanisms to be able to run the algorithms in real time~\cite{c22}. Other requirements for this work were related to a minimal number of parameters changed during locomotion process; holding the direction to the target until the target is reached and a minimum of directional changes for the energetic efficiency. This type of locomotion strategy is motivated by possible scenarios in hazardous and dangerous areas, where the robot should optimize its own behavior for a fast coverage of the targeted area~\cite{c20}.

\begin{figure}[h!]
\centering
\subfigure{\includegraphics[width=.45\textwidth]{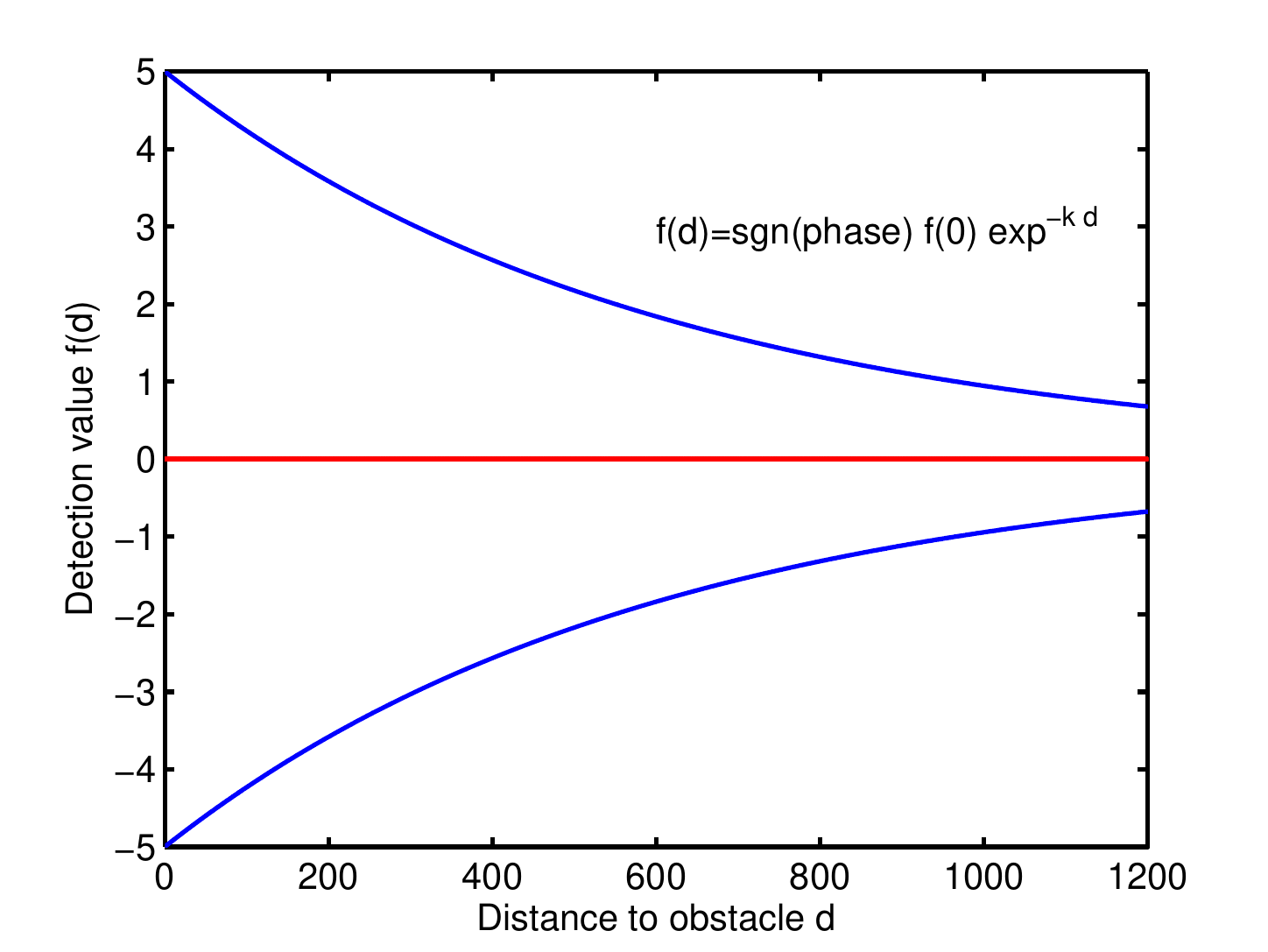}}
\caption{\it Function $f(d)=sgn(phase)f_0 e^{-k d}$, $d$ is the distance to obstacle, $k$ is the critical turning parameter.
\label{fig:Adapt}}
\end{figure}

Since a snake-like robot is not able to change rapidly the direction of movement, the gait planning and adaptation should be iteratively performed. To control a direction of motion, we use the following function $f(d)\sim\Upsilon$, see Fig.~\ref{fig:Adapt}, where $d$ is a distance to target and $sgn(phase)$ determines the branch of the function $f(d)$ to use. The value of $sgn(phase)$ indicates the turn left or right of the snake-like movement. This allows controlling the gait in a deterministic and efficient way.
The value of $f(d)$ and correspondingly $\varphi_{i}(t)$ is adapted according to the distance to the obstacle. Exponential form of the function provides smooth directional changes of the serpentine gait. Side distance sensors together with the tail sensor can be involved into the control process and additionally shape the serpentine gait. Such scenario is not inspired from the real snakes that use only the head and the skin tactile sensors but open interesting possibilities to control the shape of a snake-like robot. In situations where obstacles occur from the side these sensors shape the amplitude of the corresponding links according to the distances. In order to change the values in a smooth way, another exponential dependence between distances to the obstacles and the affected link amplitudes is used:
\begin{equation}
\widetilde{A}(d) = e^{k d}-A_{max},
\label{eq:Amplitude}
\end{equation}
where $\widetilde{A}$ is a new amplitude value calculated as a function of distance $d$, and $A_{max}$ is the maximum allowed amplitude for this kind of gait. If tail of snake detects an obstacle the angular frequency $\omega$ of the serpentine function (Eq.~\ref{eq:gate}) is also changed to avoid collision.

\section{IMPLEMENTATION}
\label{sec:Implementation}

The scenarios were first simulated in Matlab and finally implemented and tested on a real snake hardware (Fig.~\ref{fig:SnakeHardware}). In order to measure the efficiency of the algorithms, we performed two major case scenarios for locomotion; one scenario based only on head distance to the obstacles (Bio-inspired) and another scenario where all distance sensors are involved along the snake body (Tech-inspired).
\begin{figure}[h!]
\centering
\subfigure[]{\includegraphics[width=.45\textwidth]{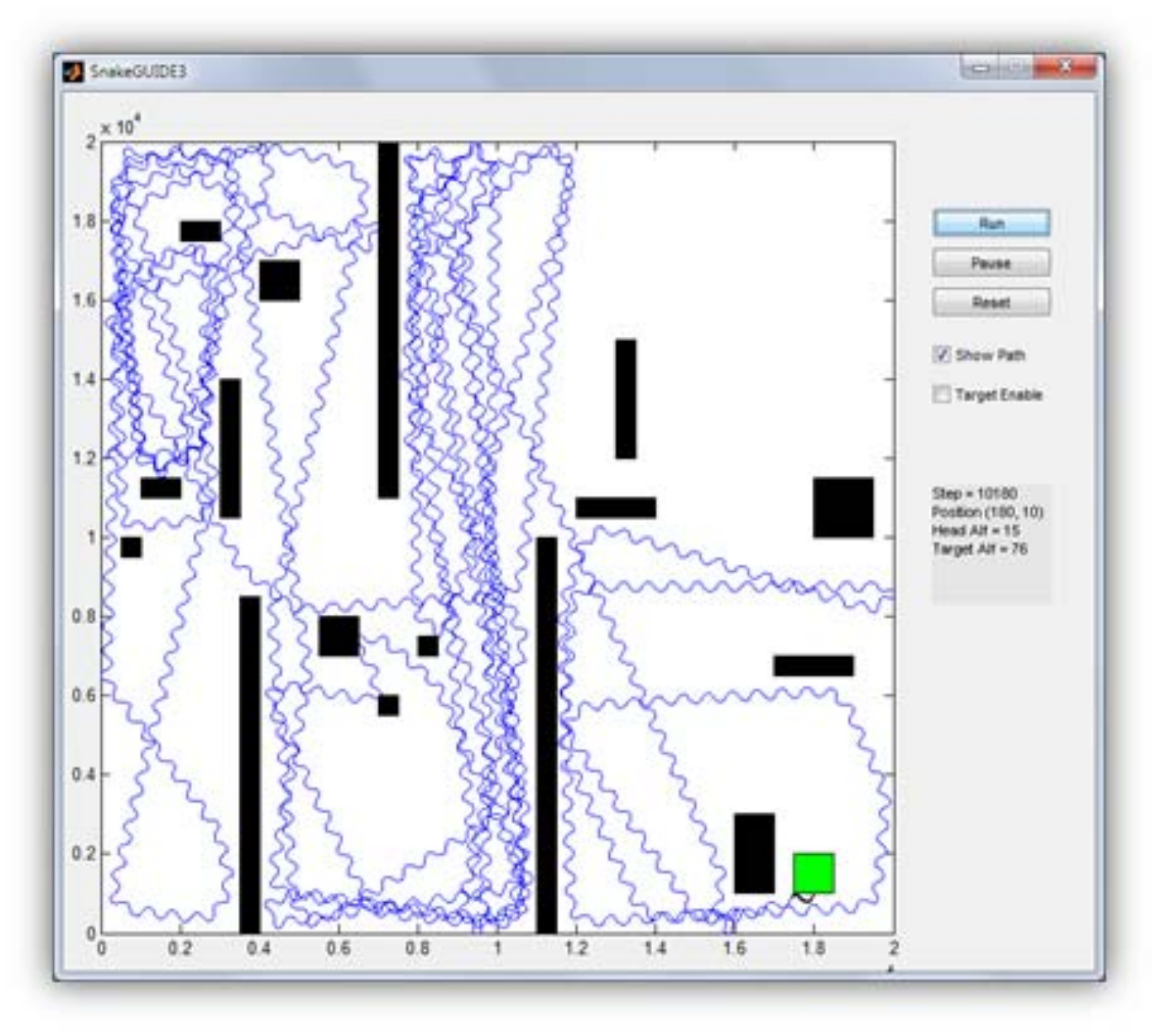}}~~~
\subfigure[]{\includegraphics[width=.45\textwidth]{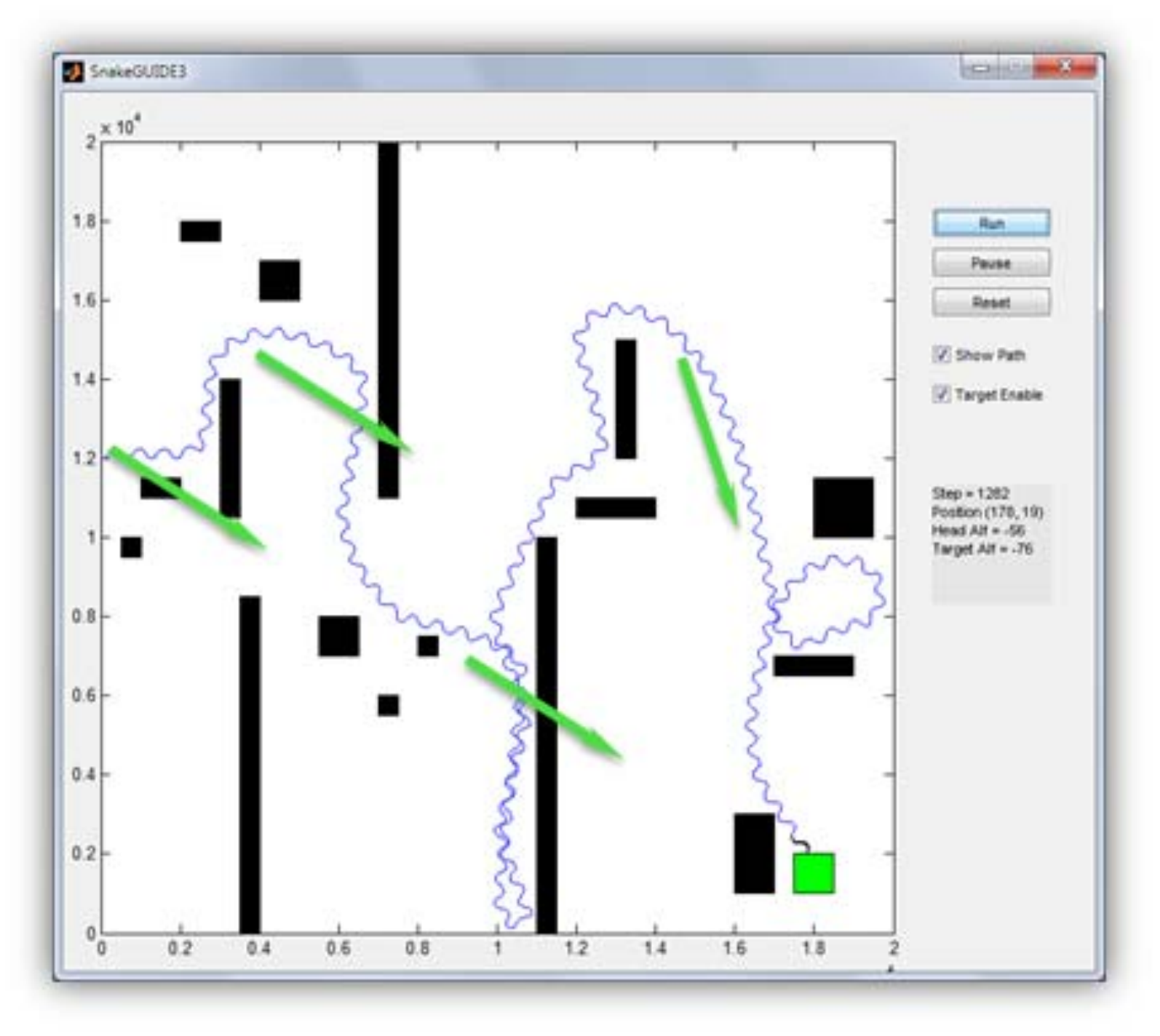}}
\caption{\it Comparison of two scenarios with and without the adaptation. \textbf{(a)} Random walk scenario until the target is reached; \textbf{(b)} adaptive target direction (green arrows) following scenario.}
\label{fig:MatlabSim}
\end{figure}

\textit{\textbf{Bio-inspired}}
In the first scenario the snake stops if it reaches the target (green box). In this case the snake fulfill only the obstacle avoidance while performing random walk, Fig.~\ref{fig:MatlabSim}(a). Even if the target is reached after a while, this kind of exploration is very inefficient and energy wasteful. In the second scenario, additionally to the obstacle avoidance, the snake tries to follow the straight direction to the target, Fig.~\ref{fig:MatlabSim}(b). On a real snake hardware the moving direction can be determined by a compass device placed on the head segment. As followed from the Fig.~\ref{fig:MatlabSim}, the second scenario provides much more effective way of reaching a target while avoiding all the obstacles.

\begin{figure}[h!]
\centering
\includegraphics[width=.6\textwidth]{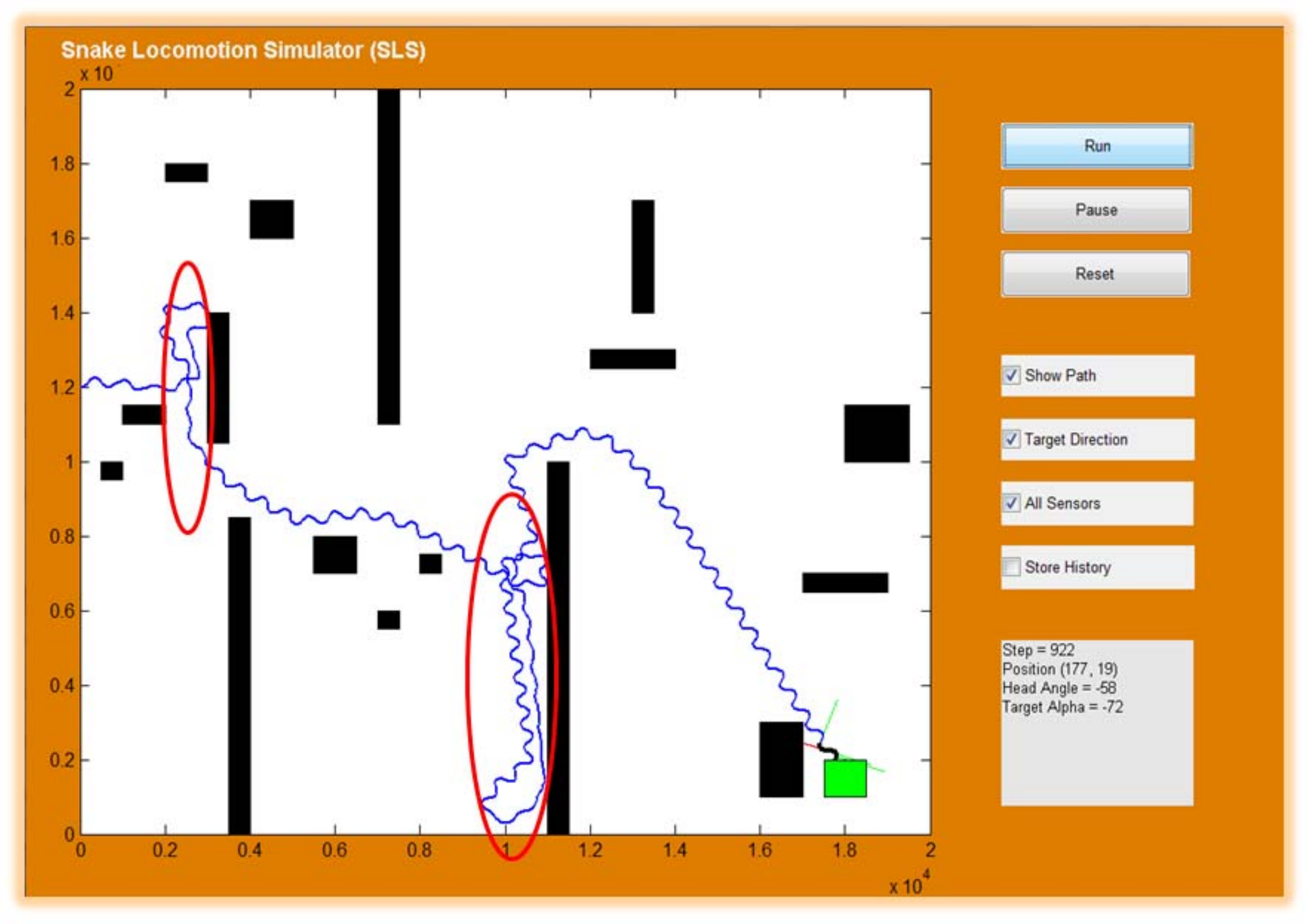}
\caption{\it Simulation of adaptive amplitude control algorithm.}
\label{fig:SimulationSensors}
\end{figure}

\textit{\textbf{Tech-inspired}}
The main difference between a real snake and a robot is the ability to percept the environment. Real snake uses different kind of sensors such as: tactile, smell, vision, acoustic or infrared radiation sensors which are more or less sophisticated dependent on the snake. Smart combination of these sensor organs allow precise foraging and navigation. For robot snake, the number and the kind of sensors are limited due to hardware capabilities. In order to extend snake robot perception we use additional IR sensors mounted around the snake. These sensors have the same function as the sensors used on the head and allow the snake to modify the shape amplitude according to the algorithm described in Sec.~\ref{sec:AdaptiveAlg}.
\begin{figure}[h!]
\centering
\subfigure{\includegraphics[width=.27\textwidth]{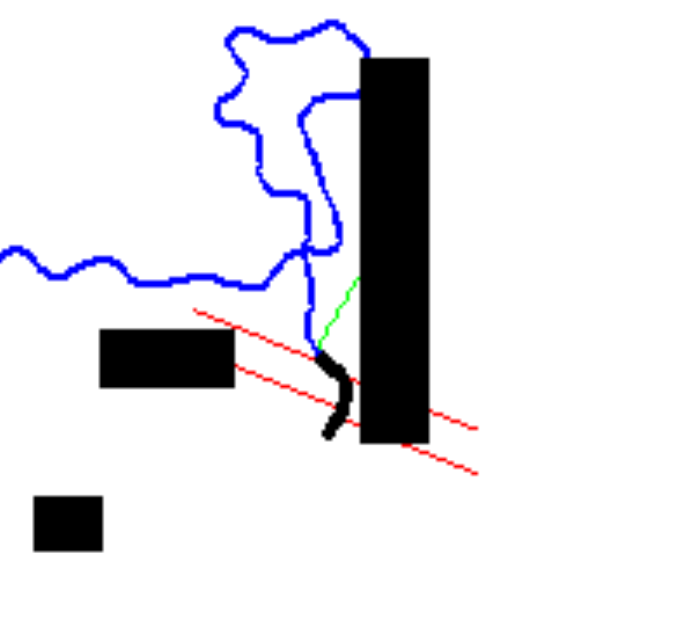}}
\subfigure{\includegraphics[width=.27\textwidth]{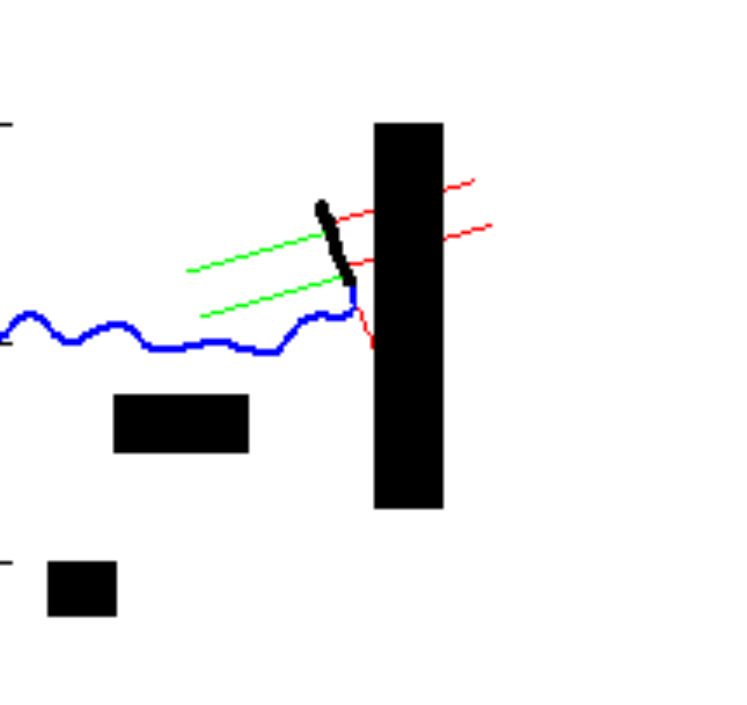}}
\subfigure{\includegraphics[width=.27\textwidth]{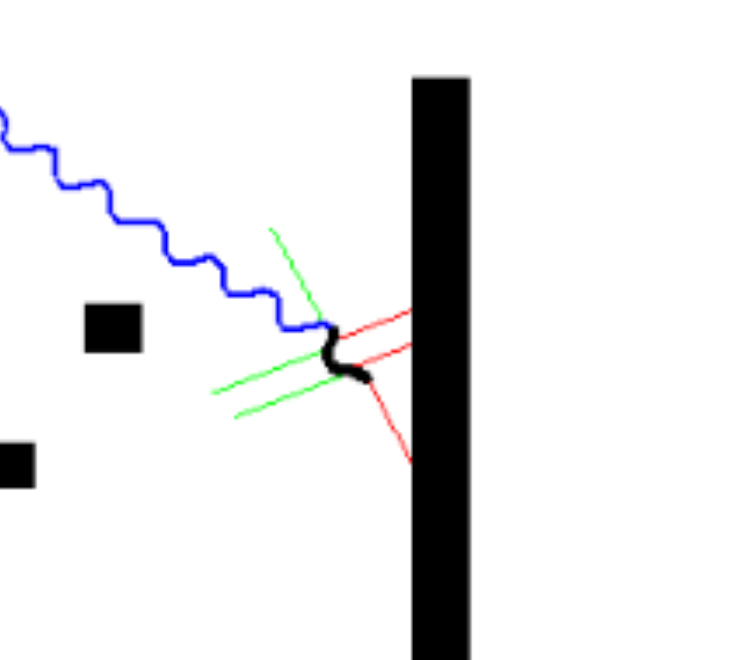}}
\caption{\it Adaptive amplitude control by using side and tail distance sensors.}
\label{fig:Sensors}
\end{figure}
Therefore, the simulation was extended with this functionality. In Figs.~\ref{fig:Sensors} and \ref{fig:SimulationSensors}, the amplitude is exponentially decreased if the snake come lateral closer to the walls or to the obstacles. The more sensors detect the obstacles the more intensive it affects the snake´s amplitude.

\section{HARDWARE IMPLEMENTATION}
\label{sec:Hardware}

\begin{figure}[h!]
\centering
\subfigure[]{\includegraphics[width=.248\textwidth]{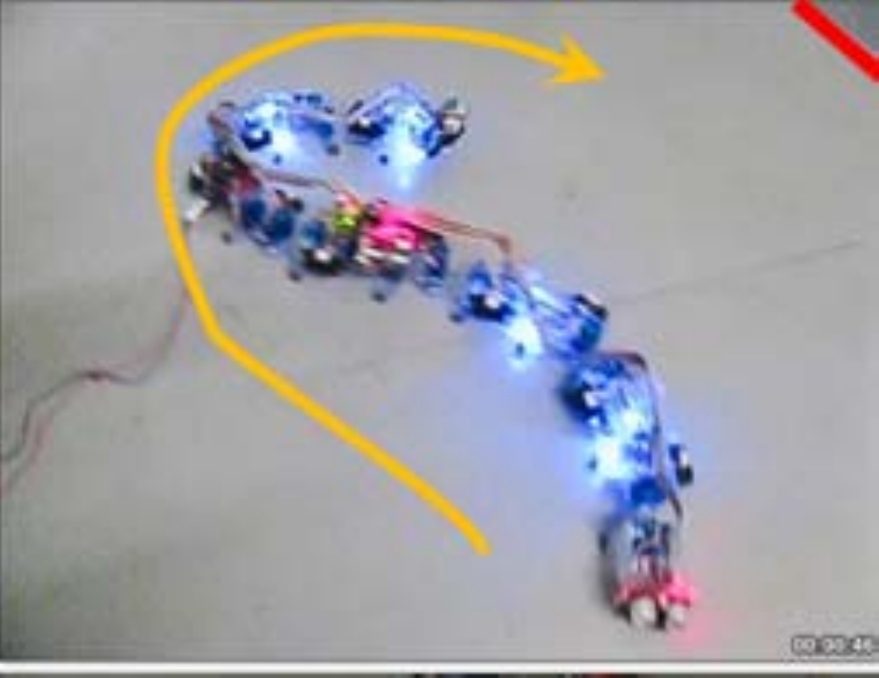}}~
\subfigure[]{\includegraphics[width=.255\textwidth]{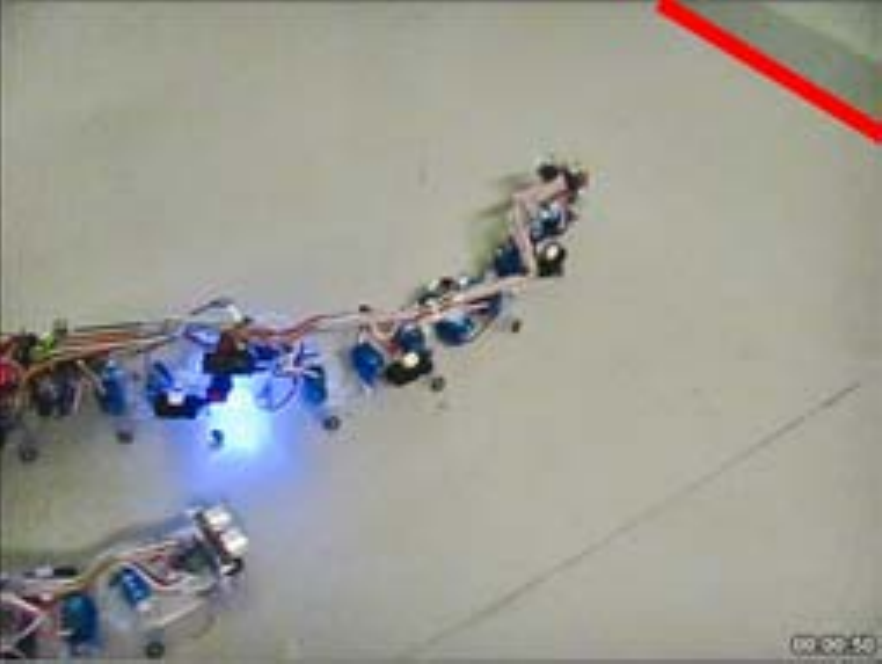}}~
\subfigure[]{\includegraphics[width=.26\textwidth]{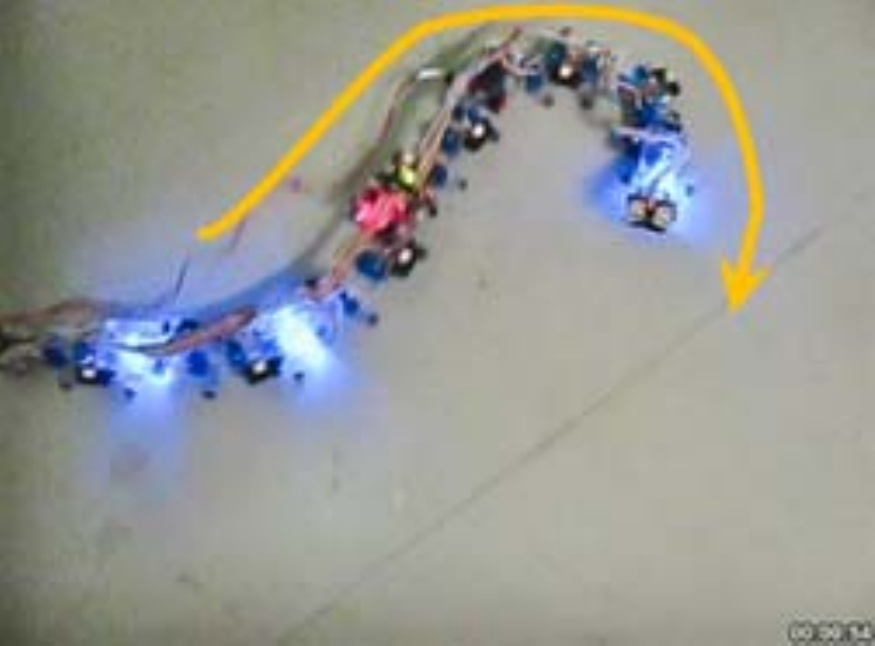}}\\
\subfigure[]{\includegraphics[width=.3\textwidth]{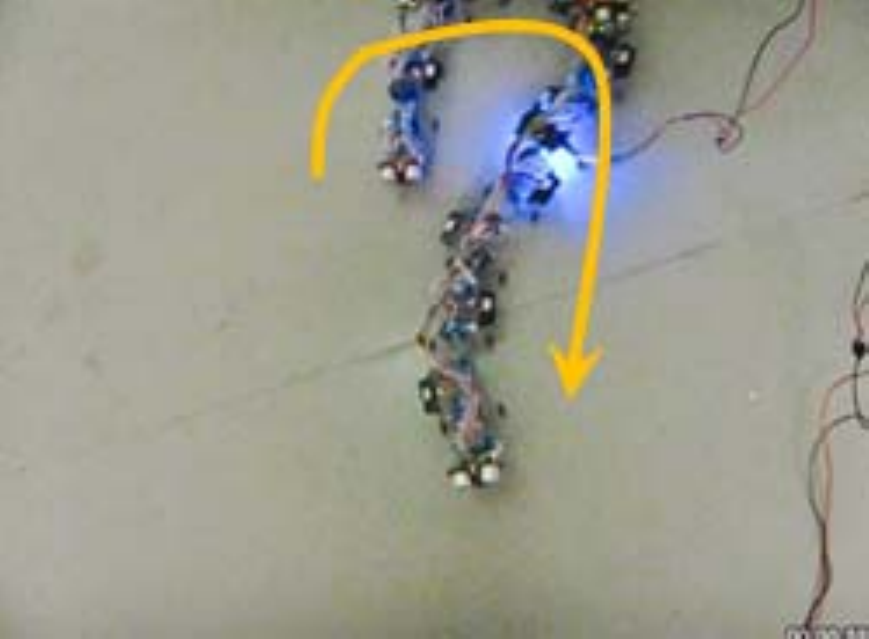}}~
\subfigure[]{\includegraphics[width=.29\textwidth]{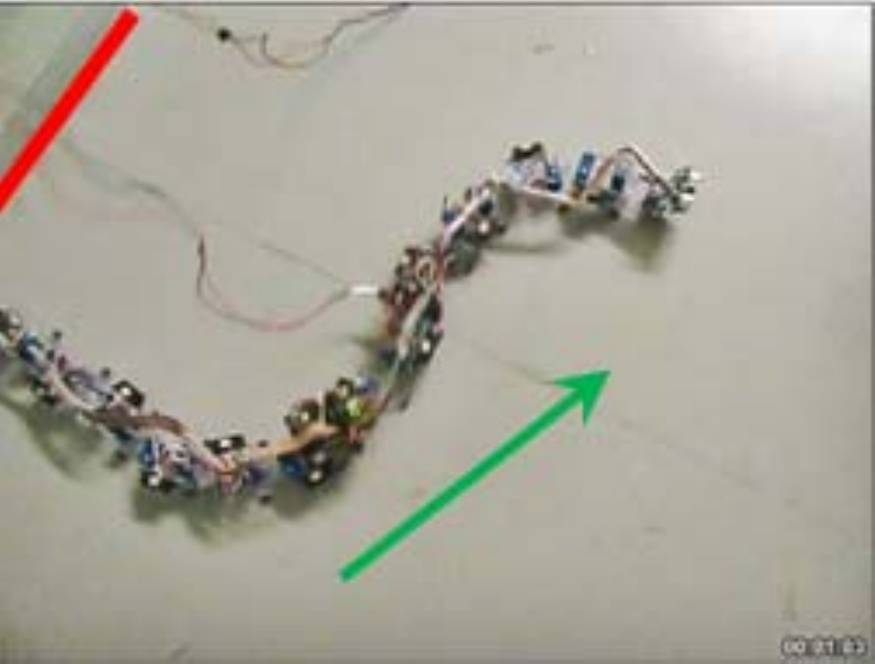}}~
\subfigure[]{\includegraphics[width=.186\textwidth]{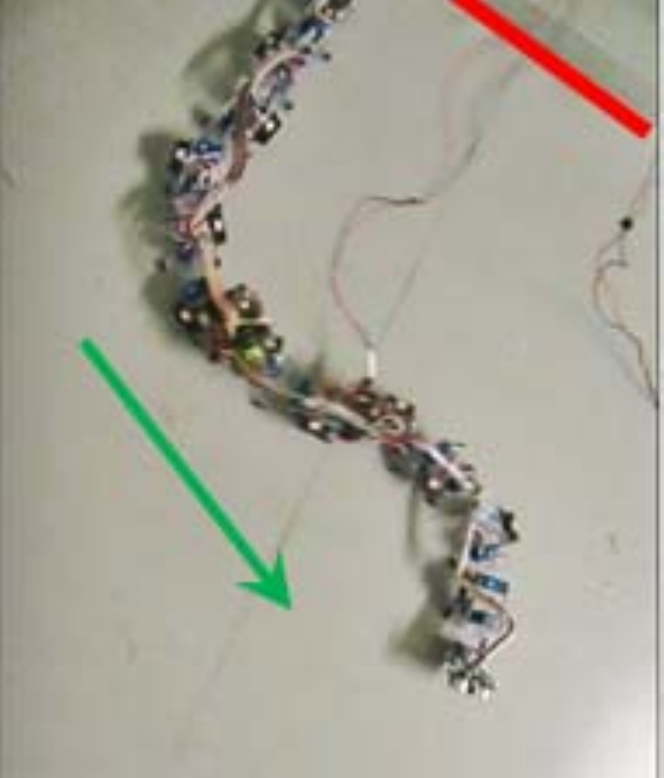}}
\caption{\it Movement of snake robot towards the goal additionally performing optical collision avoidance.}
\label{fig:Hardware}
\end{figure}

The snake hardware consists of 25 servo motors assembled alternately in horizontal and vertical directions. Therefore, the snake can perform planar movement as well as limited spatial locomotion. The snake is controlled by a Programmable System on Chip (ARM Cortex-M3 PSoC 5) microprocessor from Cypress, which directly generates PWM signals for servo motors. All distance sensors are directly connected to ADC of PSoC chip; compass, sonar and accelerometer are accessible through I2C bus. The first real scenarios demonstrates the collision avoidance by using only the head distance sensor (Fig.~\ref{fig:Hardware}, (a)-(f)). The range of the IR distance sensors is approximately one and a half meter. The servo motor position is limited to 30 degrees to fulfill a whole period cycle ($2\pi/12$) for locomotion in 2D.

\section{3D LOCOMOTION}
\label{sec:3DLocomotion}

\begin{figure}[h!]
\centering
\subfigure{\includegraphics[width=.3\textwidth]{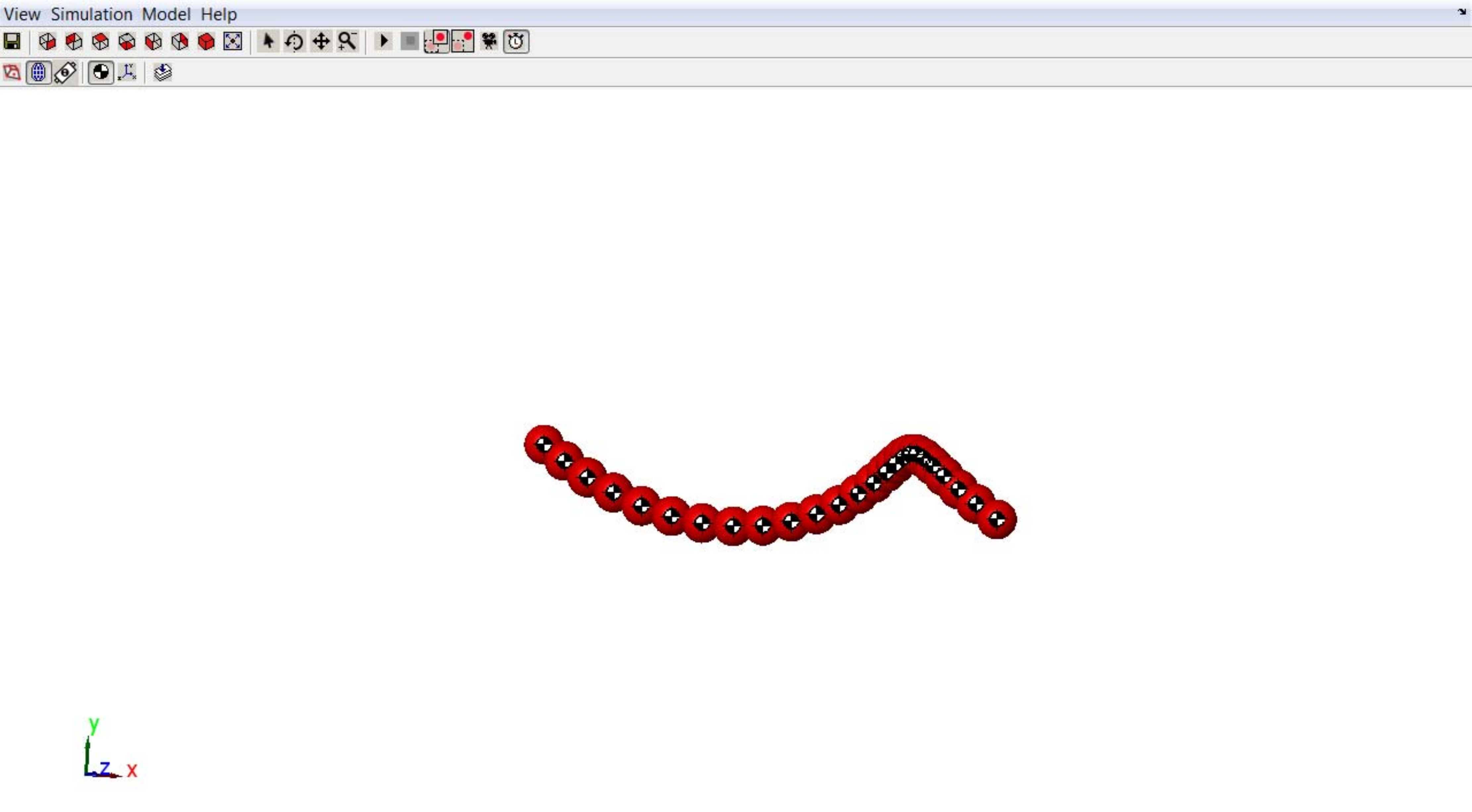}}
\subfigure{\includegraphics[width=.3\textwidth]{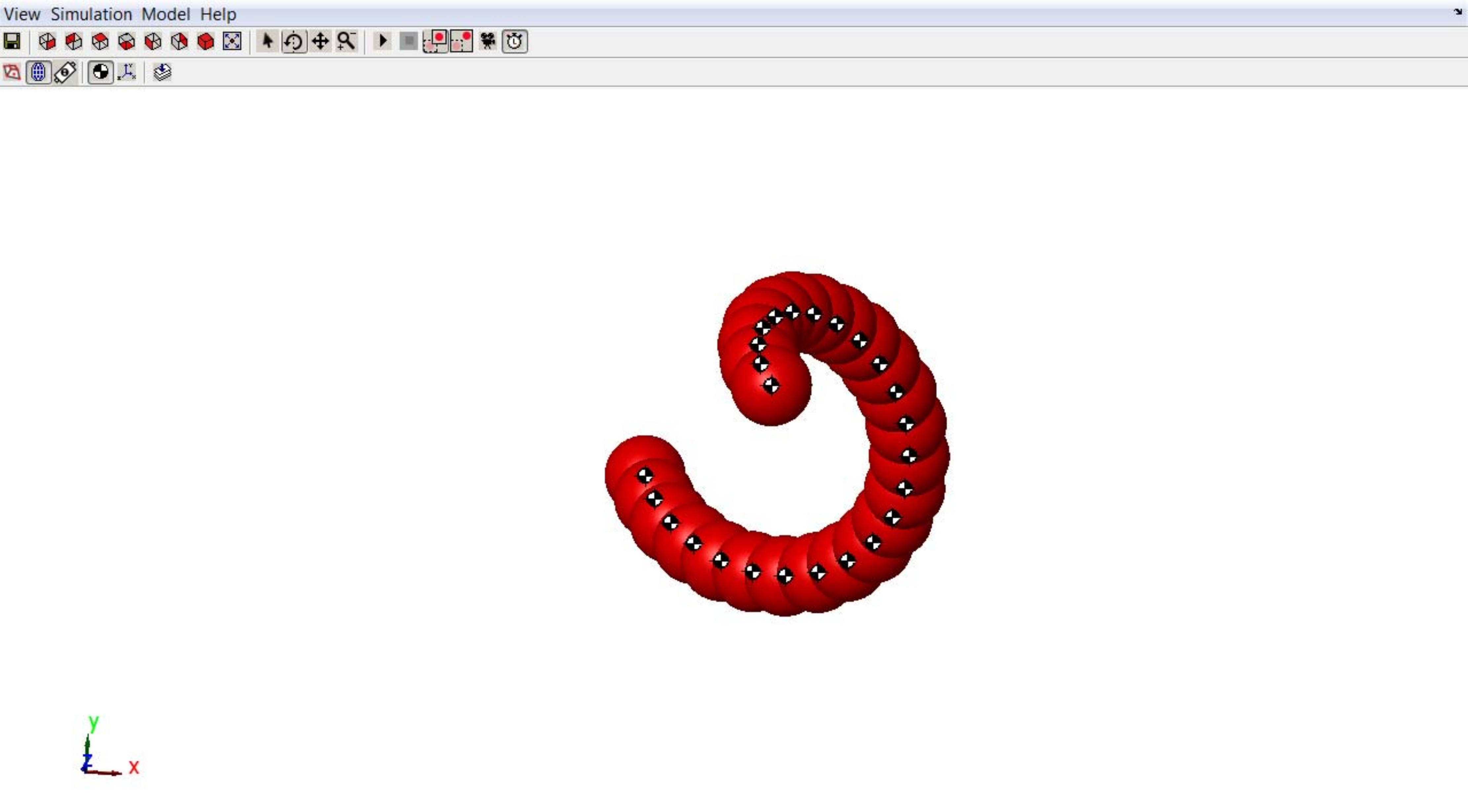}}
\subfigure{\includegraphics[width=.3\textwidth]{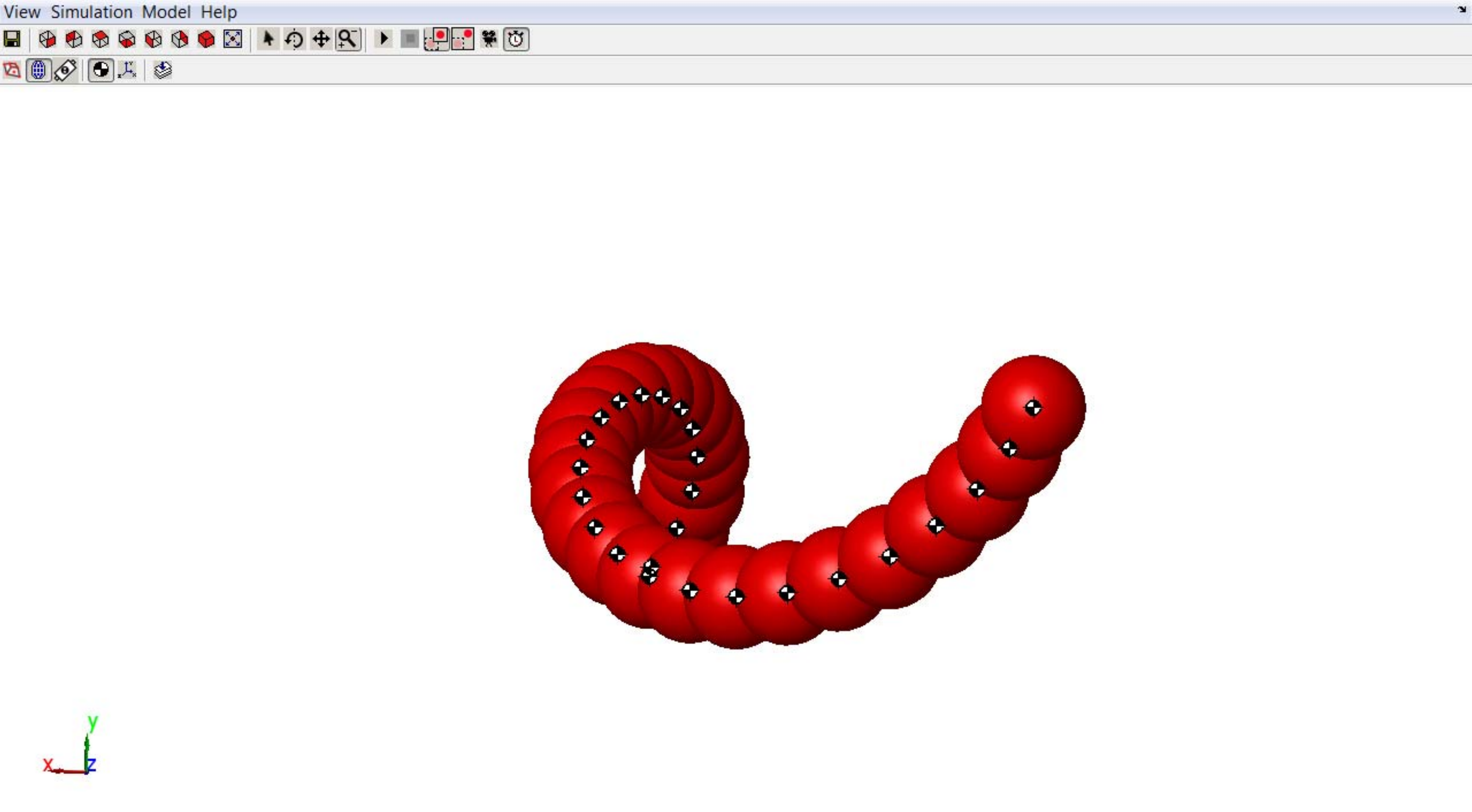}}
\caption{\it Investigation of 3D gaits in Matlab SimMechanics without considerations of contact and friction forces.}
\label{fig:3dGates}
\end{figure}

The adaptive algorithm for collision avoidance explained in Sec.~\ref{sec:AdaptiveAlg} is at the current state applicable only for 2D locomotion. In order to transfer this idea to a 3D case scenario we need to investigate gaits that allow to move the snake in a 3D environment. In \cite{Transeth}, a good overview of different possible gait formulations and locomotion strategies in planar as well as for 3D environment is summarized. One of the promised gaits for a 3D locomotion is the combination of lateral undulation gait with the additional sinusoidal generator that controls the vertical gait. Therefore, additional parameter $\phi_0$ is required that control the phase difference between the horizontal and vertical gait:
\begin{equation}
\varphi_{i,v}(t) = A_{v} sin(\omega_{v} t + (i-1)\phi_{v} + \phi_{0}) + \Upsilon_{v}, ~~~(i=1,...,n-1),
\label{eq:gate3D}
\end{equation}
where index $v$ means the vertical direction. Applying only the vertical gait the snake will perform a caterpillar-like gait. Combination of both creates a twist-like movement patterns like shown in Fig.~\ref{fig:3dGates}.

\section{SNAKE DYNAMICS}
\label{sec:Dynamics}

For studying the dynamics of the snake-like topology, we used Matlab Simulink with the combination of the SimMechanics toolbox. SimMechanics uses the second order Euler-Lagrange dynamic equations for calculation of dynamics \cite{Wood} and allow efficient and fast multi-body simulation.
\begin{figure}[h!]
\centering
\includegraphics[width=1.\textwidth]{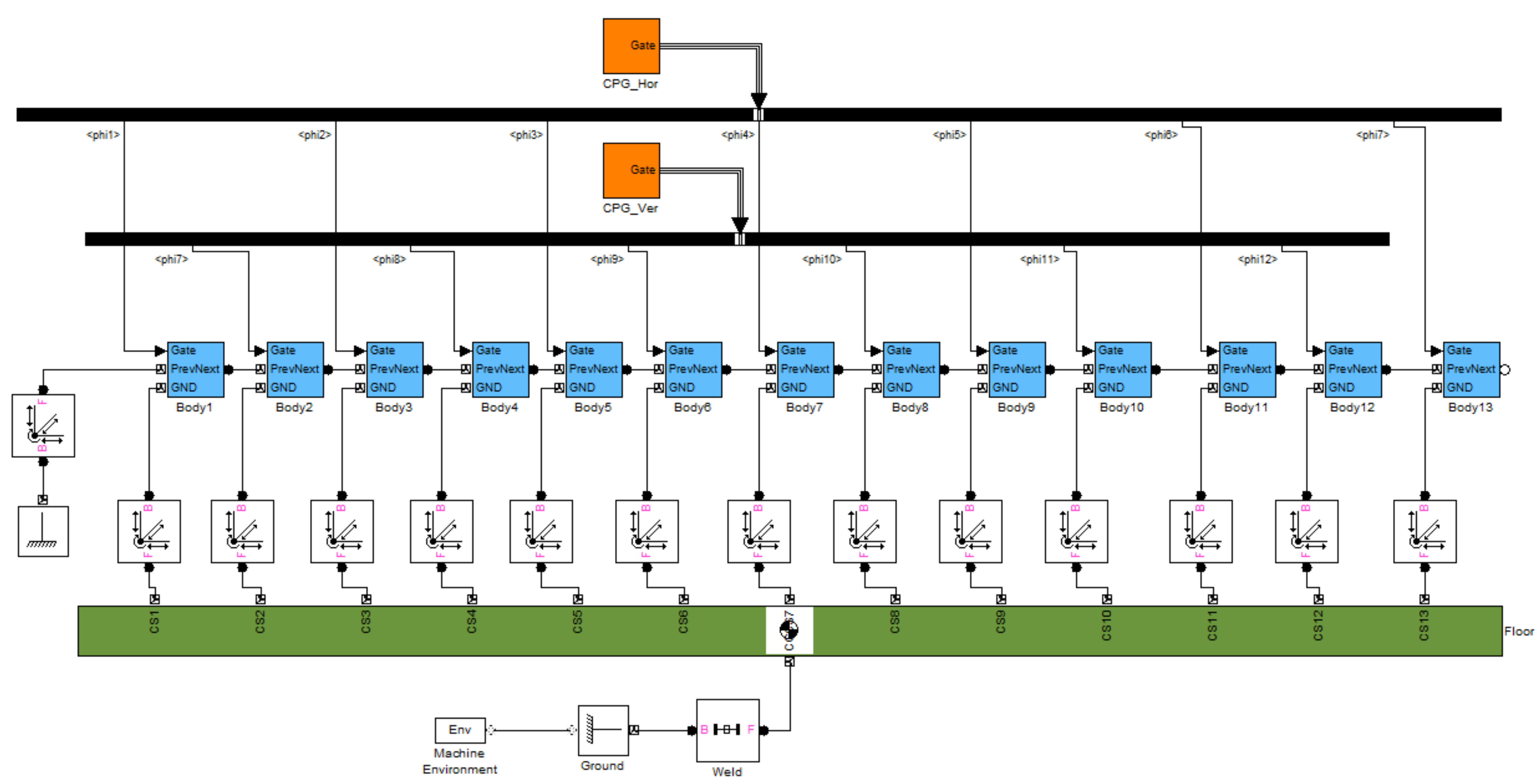}
\caption{\it Simulink model for 3D snake with 13 links arranged alternately along z and x axes.
\label{fig:SimulinkModel}}
\end{figure}
The snake robot is modelled following the idea that modules are connected to the ground by six DOF joints and hence allow to translate and rotate each module along all six axis in space (Fig.~\ref{fig:SimulinkModel}). This arrangement is necessary to simulate friction and body contact with the ground. A spring-damper system (Eq.~\ref{eq:NormalForce}) is used to model the normal forces during the ground contacts. To simulate the hardness of the ground the spring coefficient $k$ is set to a rather high value. The damping coefficient $d$ is smaller in contrast to $k$ and is necessary to damp out the oscillations.
\begin{equation}
F^{i}_{n} = -k p^{i}_{z} - d v^{i}_{z}  ~~~~~~~~\forall p^{i}_{z} < 0,
\label{eq:NormalForce}
\end{equation}
where $i$ denote the link number and $p^{i}_{z}$ is the position and $v^{i}_{z}$ is the velocity component in $z$-direction. On a real snake, we use wheels on each module in order to mimic the friction forces. As pointed out by \cite{Saito}, the friction of real snakes is much higher in the direction tangential to snake body length than the friction in the forward direction of the body. It is given through the skin shape of real snakes and reduce the side slipping.

\begin{figure}[h!]
\centering
\subfigure[]{\includegraphics[width=.5\textwidth]{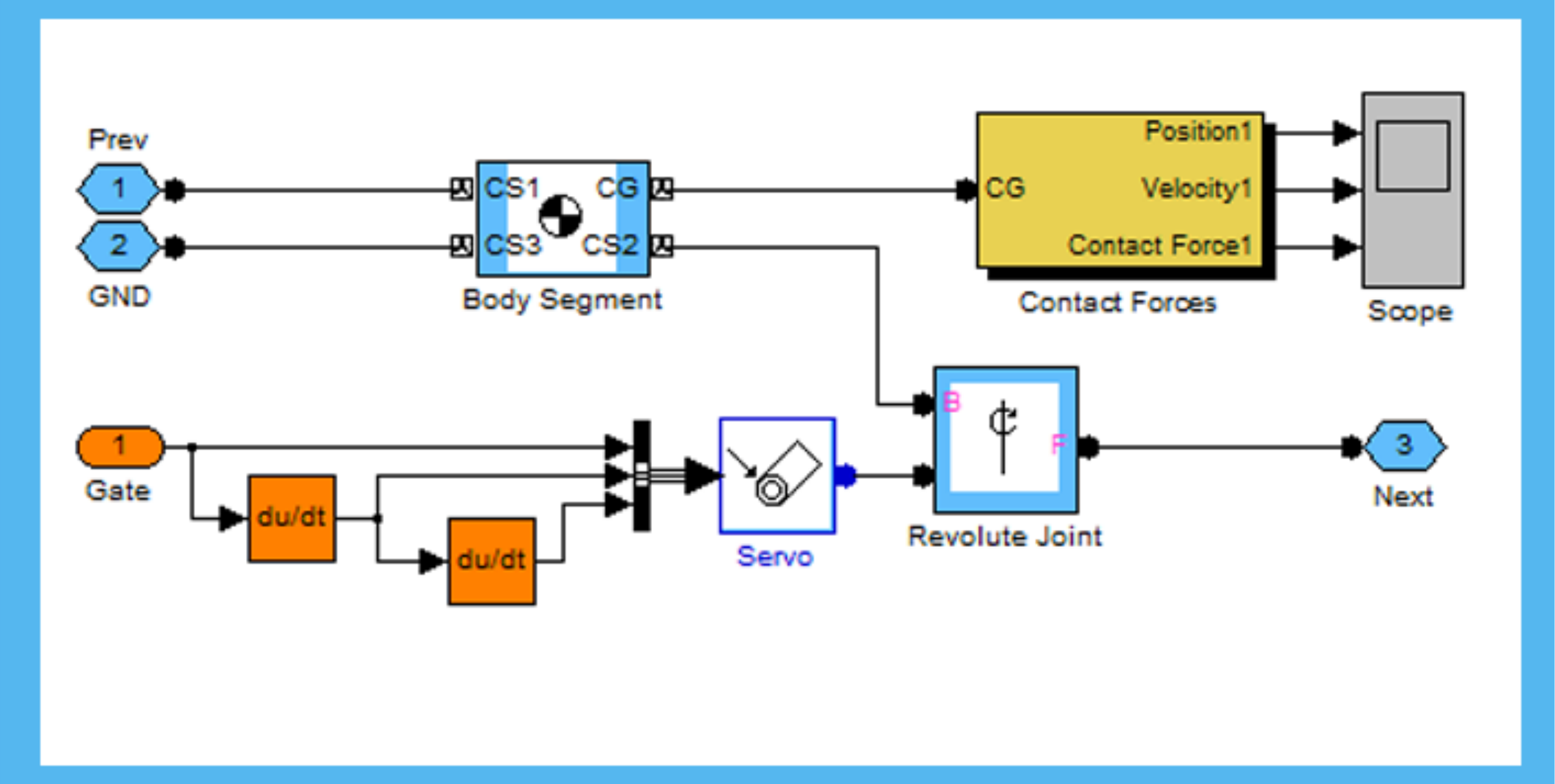}}
\subfigure[]{\includegraphics[width=.464\textwidth]{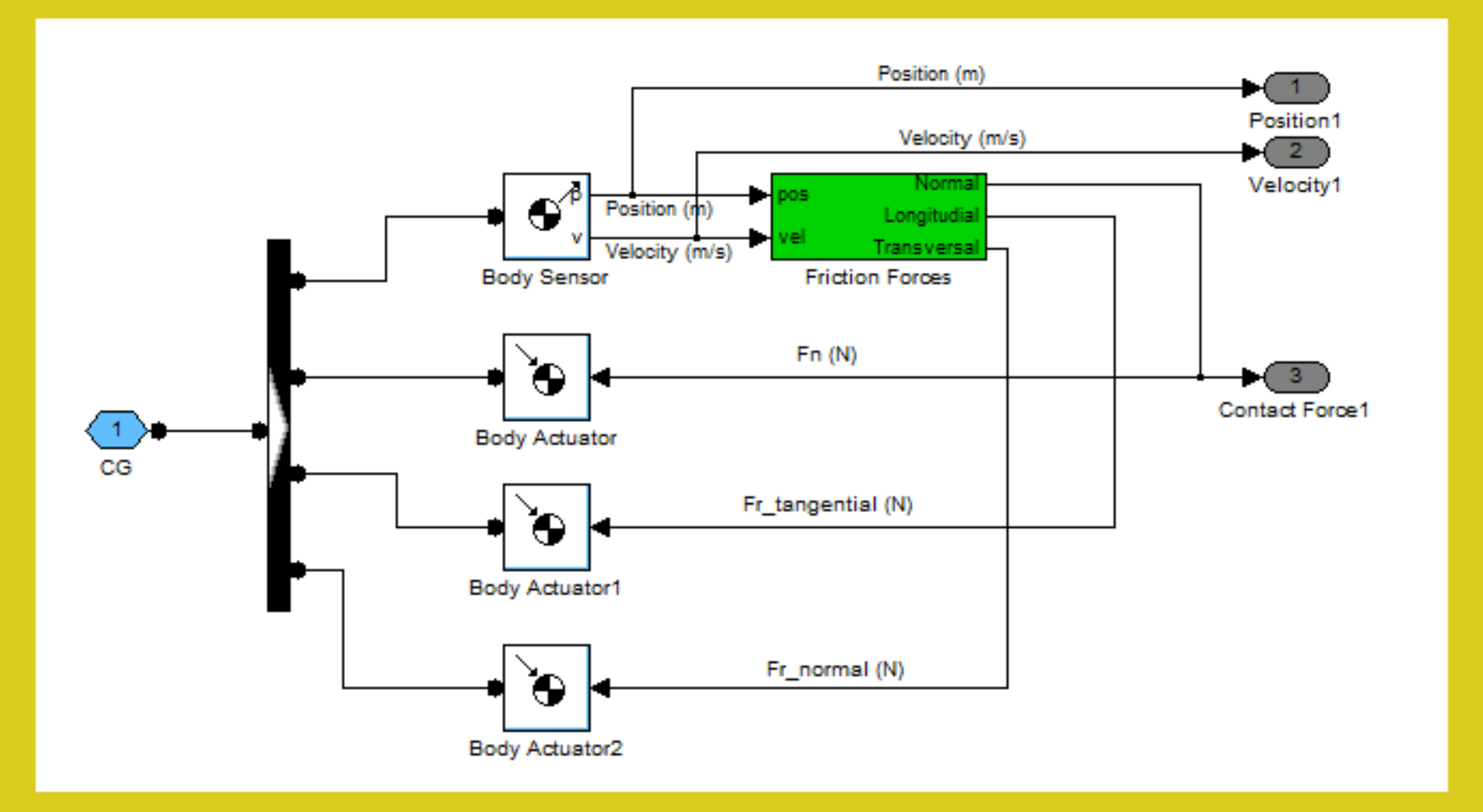}}
\subfigure[]{\includegraphics[width=.58\textwidth]{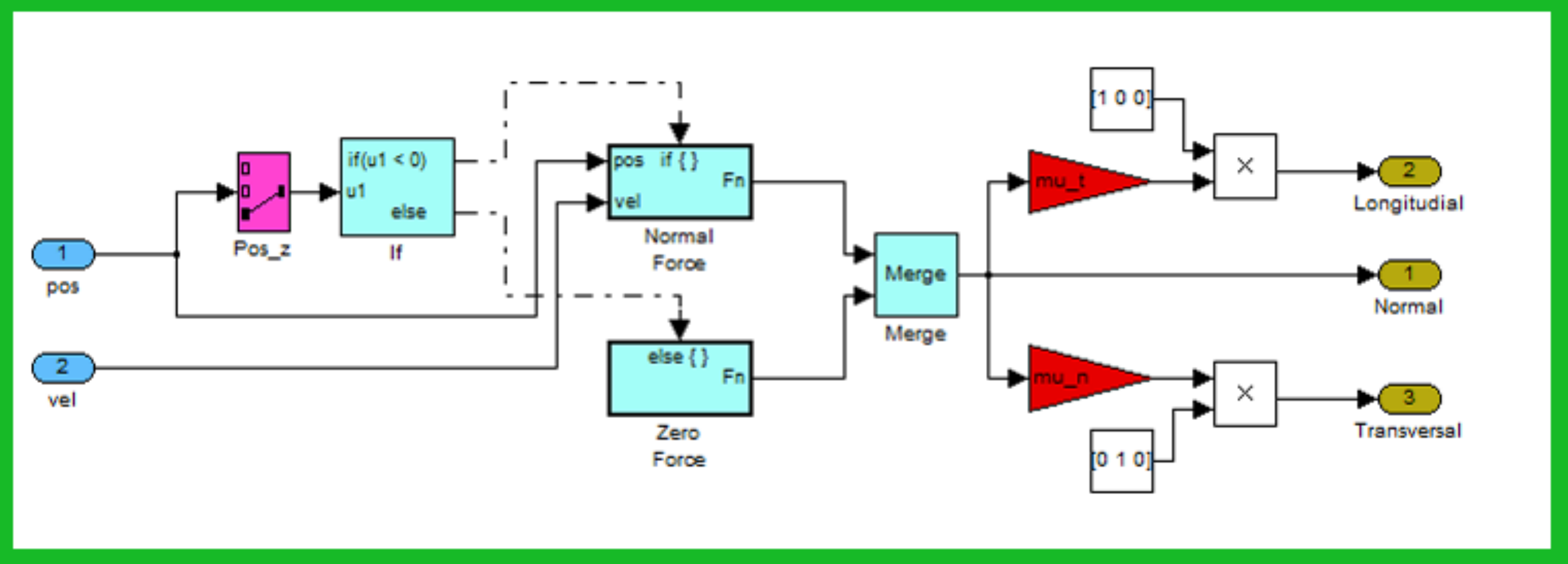}}
\subfigure[]{\includegraphics[width=.358\textwidth]{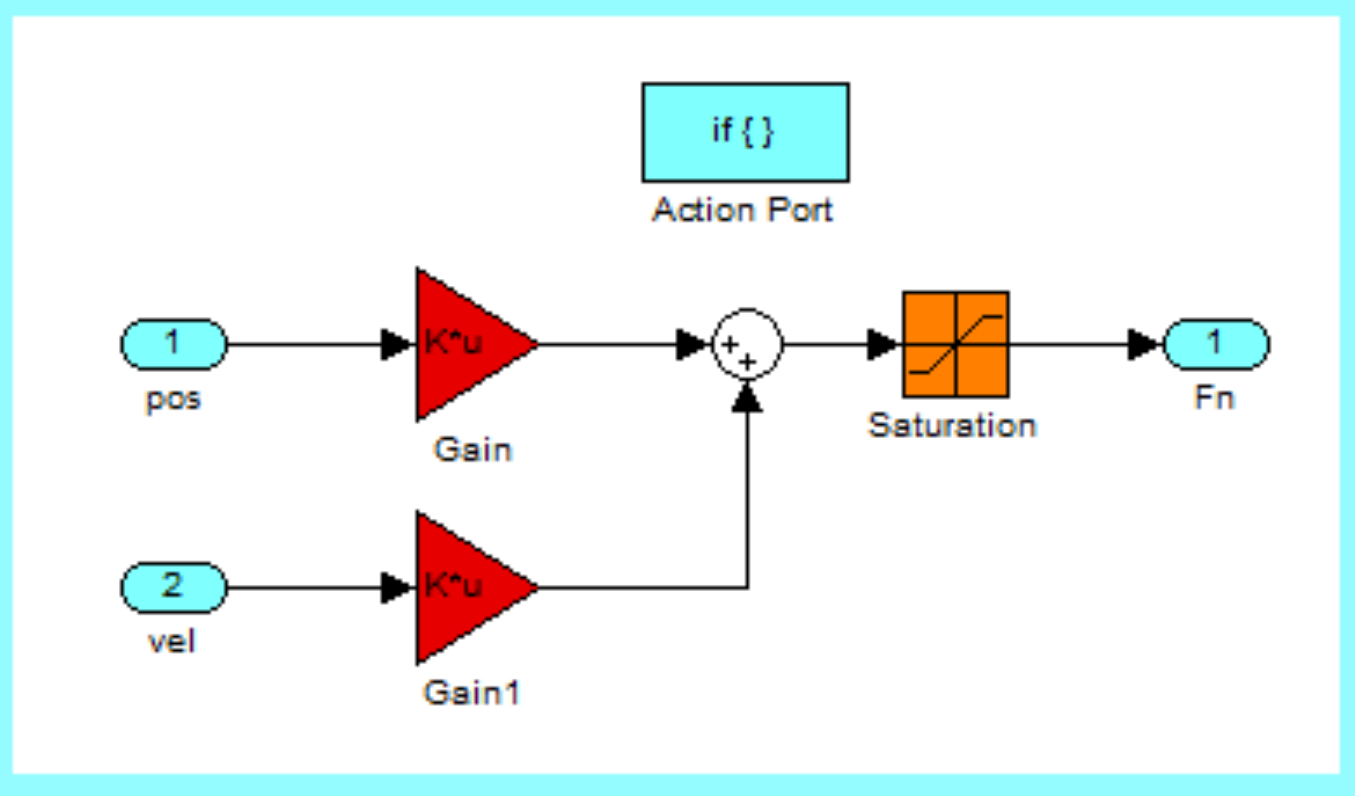}}
\caption{\it Cascaded model of snake in Simulink through structural levels marked by colored border ; \textbf{(a)} Structure of a single link element (Level 1), \textbf{(b)} General contact forces block (Level 2), \textbf{(c)} Normal, longitudinal and transversal force model (Level 3), \textbf{(d)} Normal force (Level 4).}
\label{fig:Friction}
\end{figure}

To model these frictions and simultaneously not to overload the simulation, we use the most simple friction model in both planar directions:~$F^{i}_{R_{l}} = \mu_t F^{i}_{n}$ ~and ~$F^{i}_{R_{t}} = \mu_n F^{i}_{n}$, where indexes $l$ and $t$ denote longitudinal (tangential) and transversal (normal) directions. We do not consider viscose friction or slip-stick effects because the mass and the speed of the snake links is quite low and hence these parts can be neglected at the current state. Figure~\ref{fig:Friction} shows a cascaded Simulink model, where colored borders show hierarchical structured contents of the corresponding blocks.

\begin{figure}[h!]
\centering
\includegraphics[width=.99\textwidth]{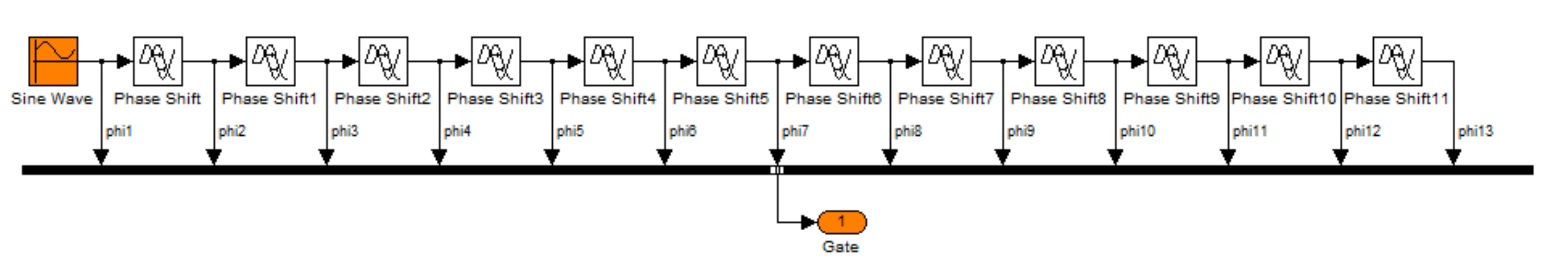}
\caption{\it Serpentine gait implementation in Simulink using delay function blocks.}
\label{Fig:CPG}
\end{figure}

The gait for the serpentine locomotion from Eq.~\ref{eq:gate} can be implemented by providing delay function blocks (see Fig.~\ref{Fig:CPG}) and simulate phase shift between each link.

\begin{figure}[h!]
\centering
\subfigure{\includegraphics[width=.8\textwidth]{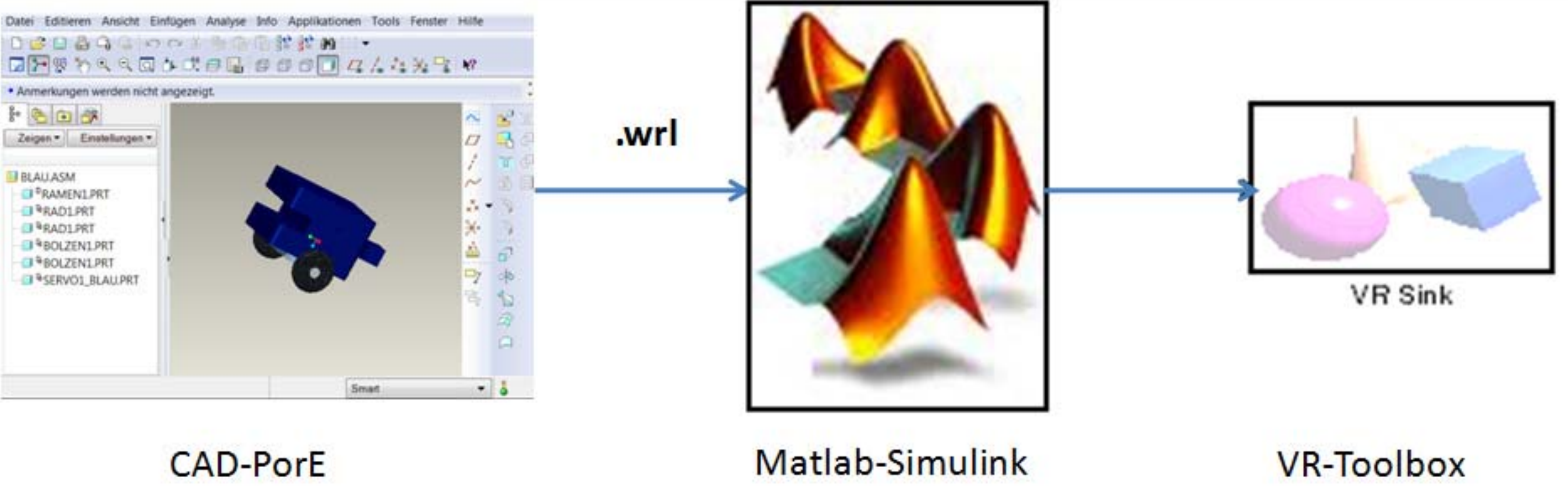}}
\caption{\it Visualization strategy of snake locomotion in VR.}
\label{fig:VR}
\end{figure}

The scenarios are also implemented in the virtual reality (VR) environment by using the VR toolbox, which allows more realistic 3D visualization of our Simulink models. This framework enables the modeling of snake mechanics in a CAD software such as Pro Engineer and transporting the data to the VR toolbox, see Fig.~\ref{fig:VR}.

\begin{figure}[ht]
\centering
\subfigure[]{\includegraphics[width=.235\textwidth]{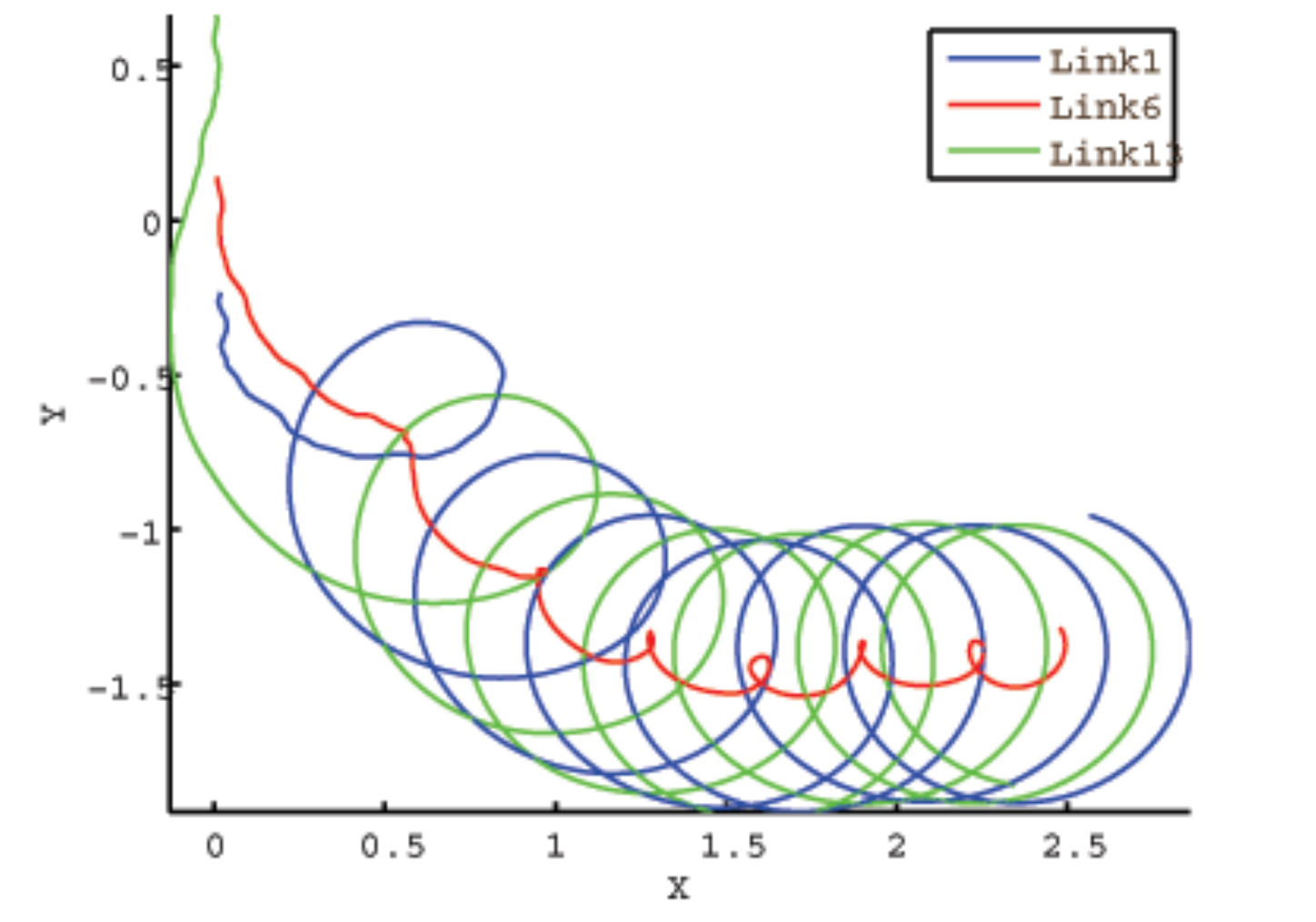}}
\subfigure[]{\includegraphics[width=.235\textwidth]{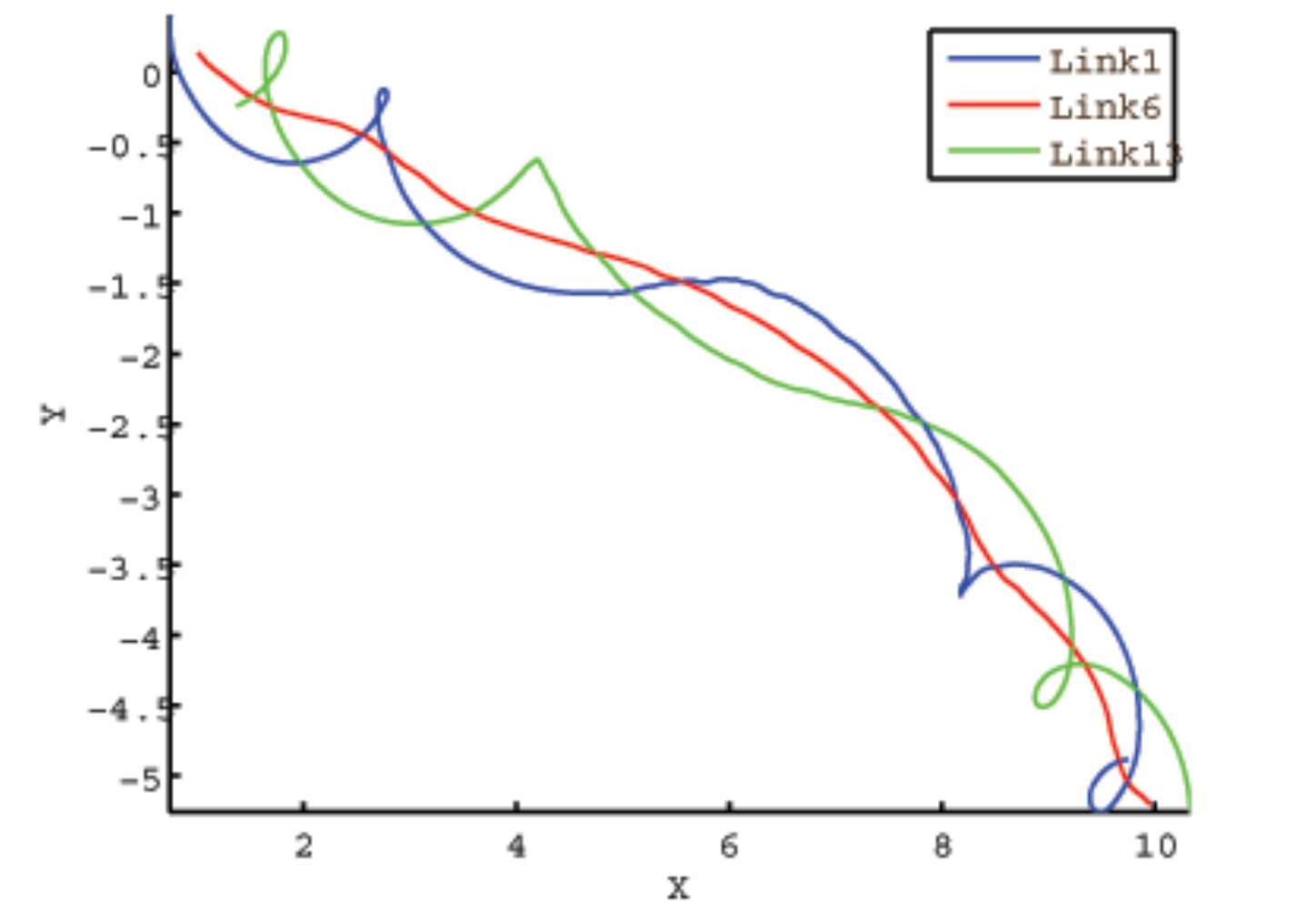}}
\subfigure[]{\includegraphics[width=.235\textwidth]{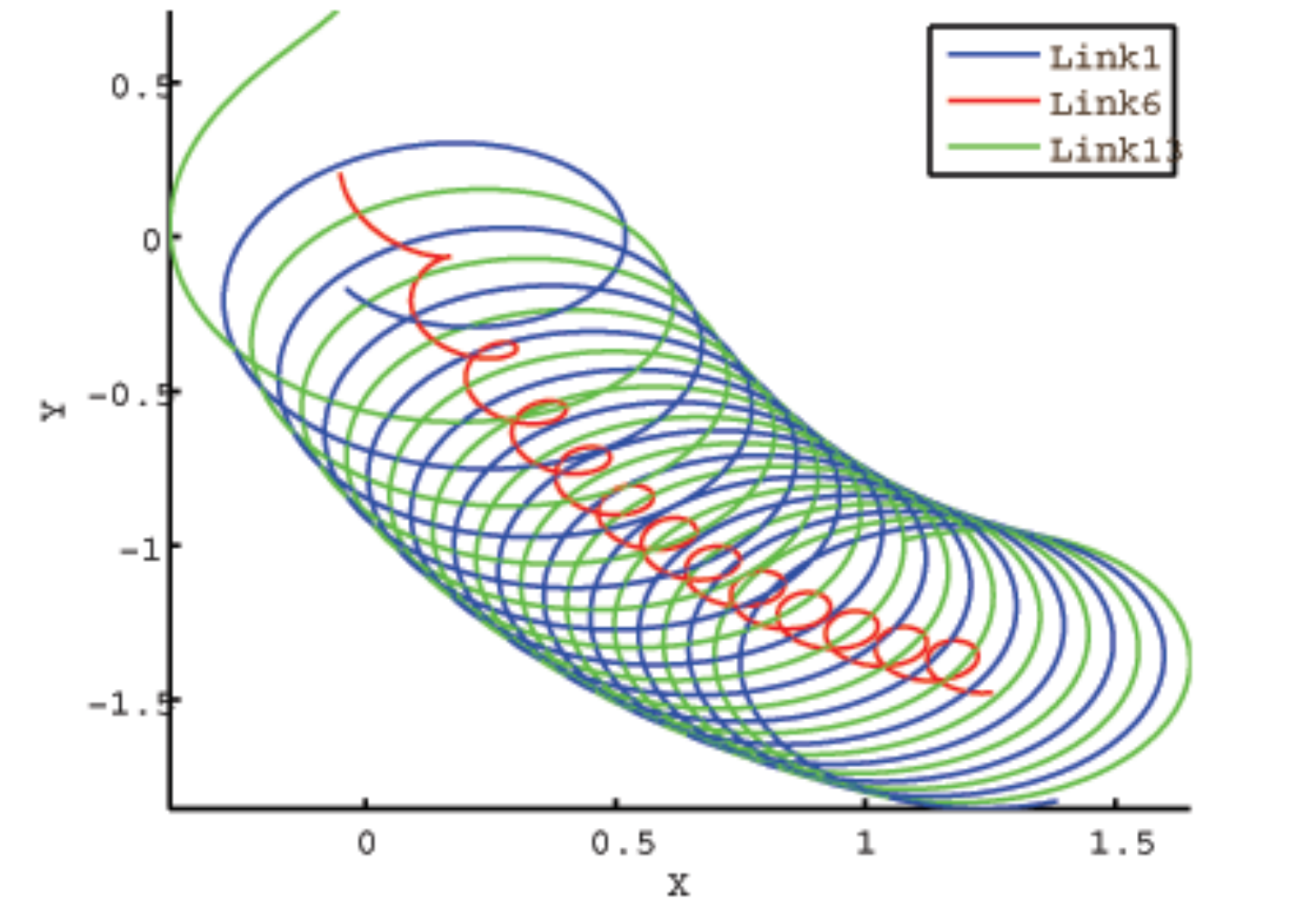}}
\subfigure[]{\includegraphics[width=.235\textwidth]{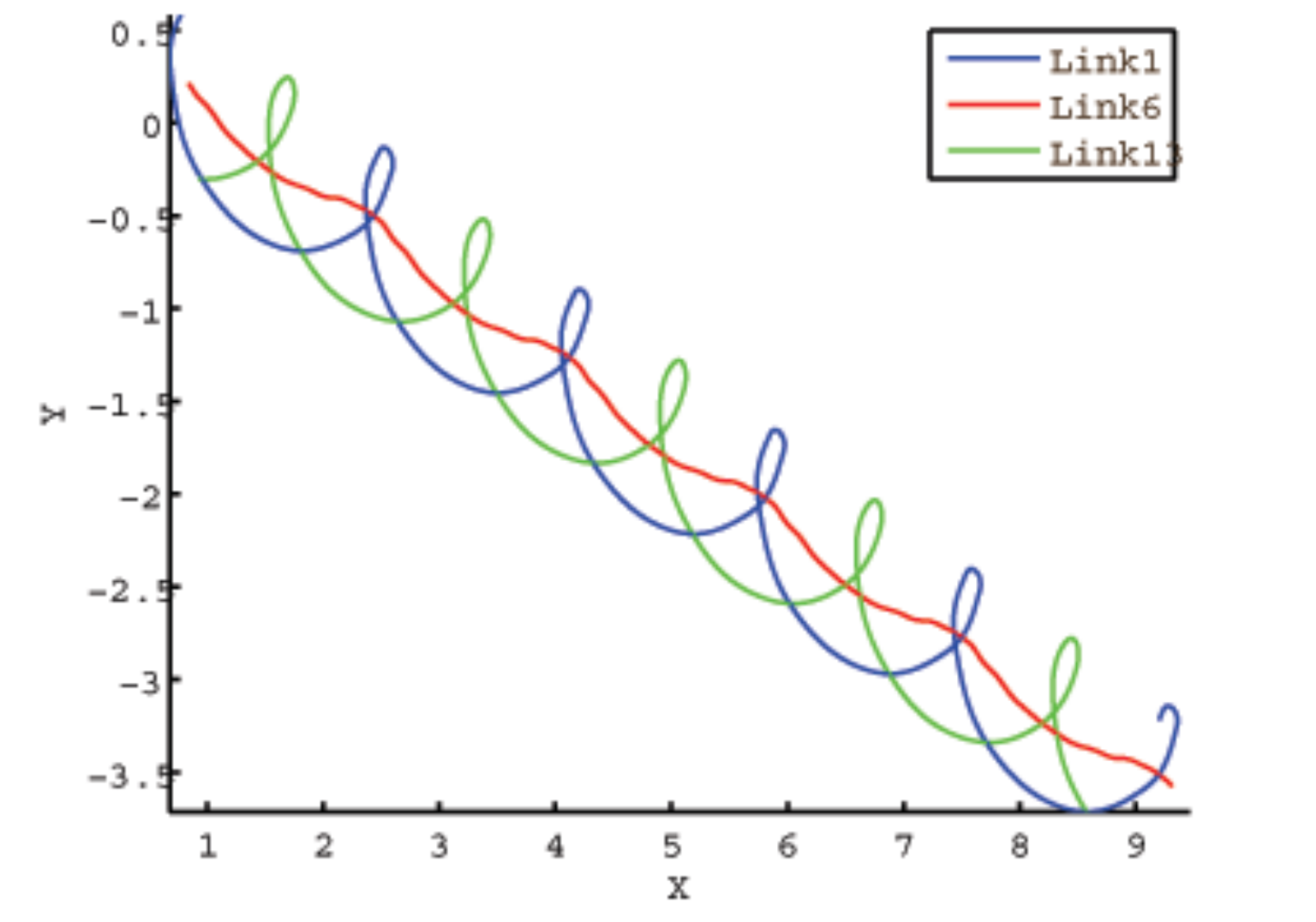}}\\
\subfigure[]{\includegraphics[width=.235\textwidth]{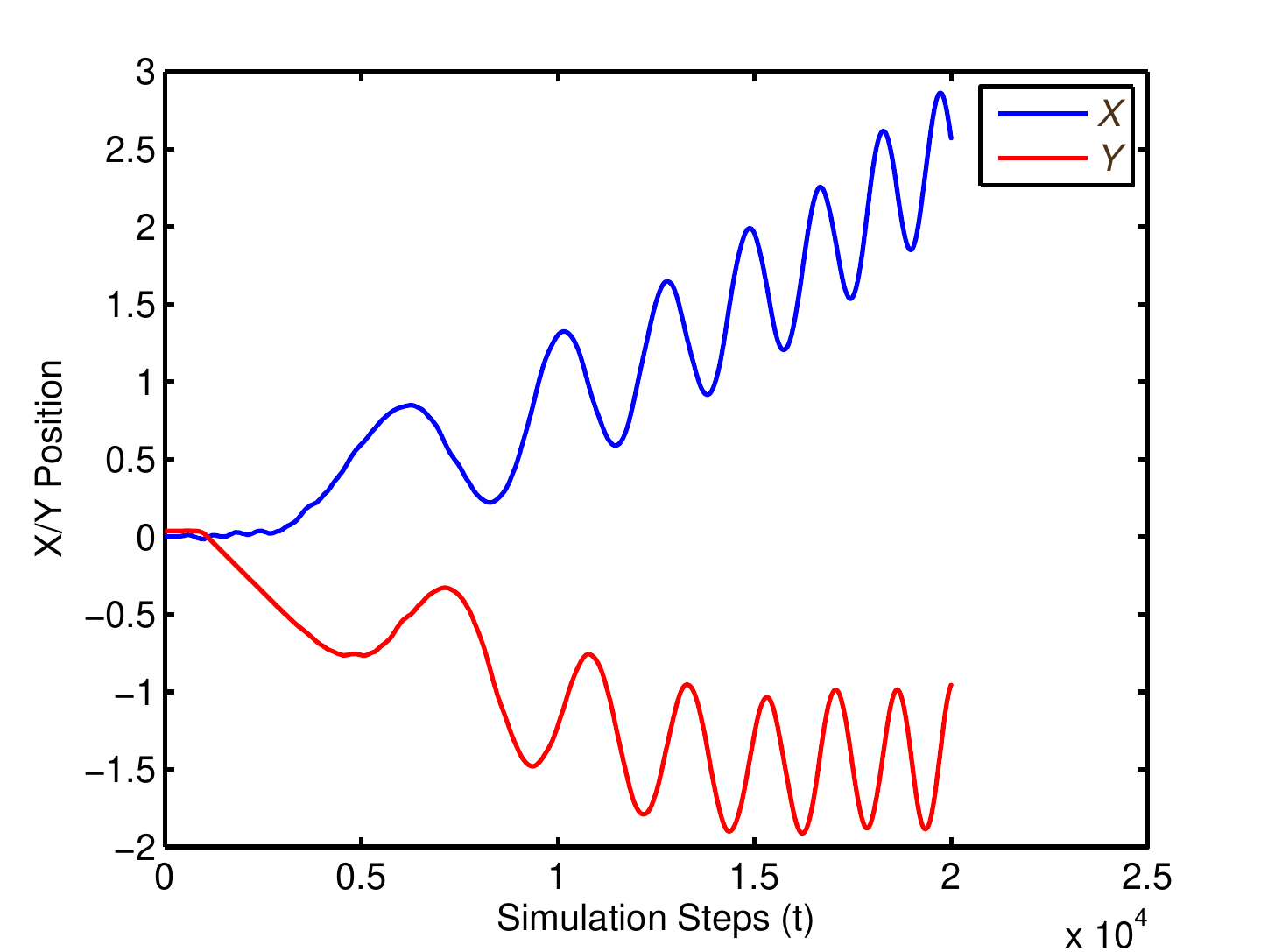}}
\subfigure[]{\includegraphics[width=.235\textwidth]{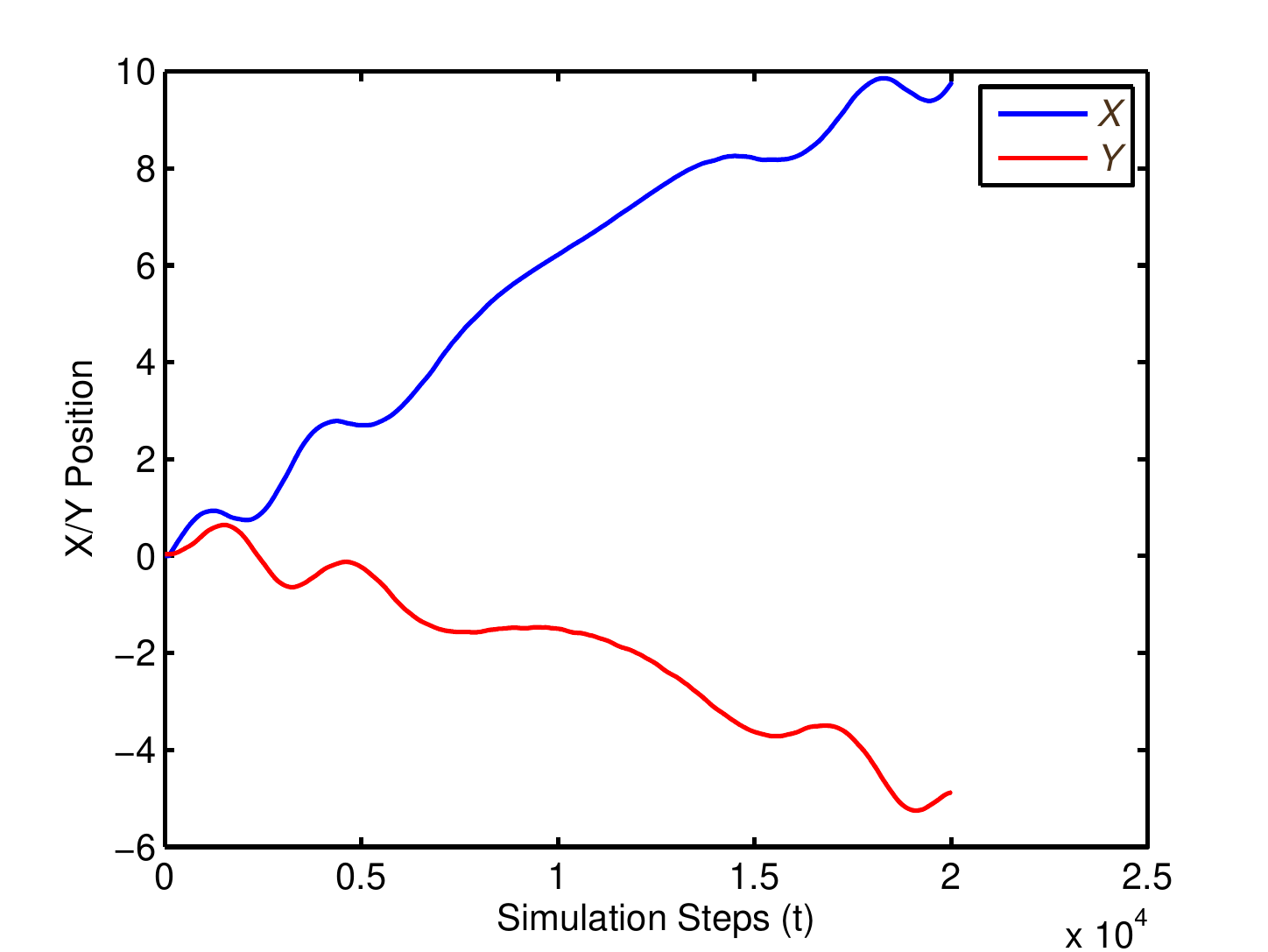}}
\subfigure[]{\includegraphics[width=.235\textwidth]{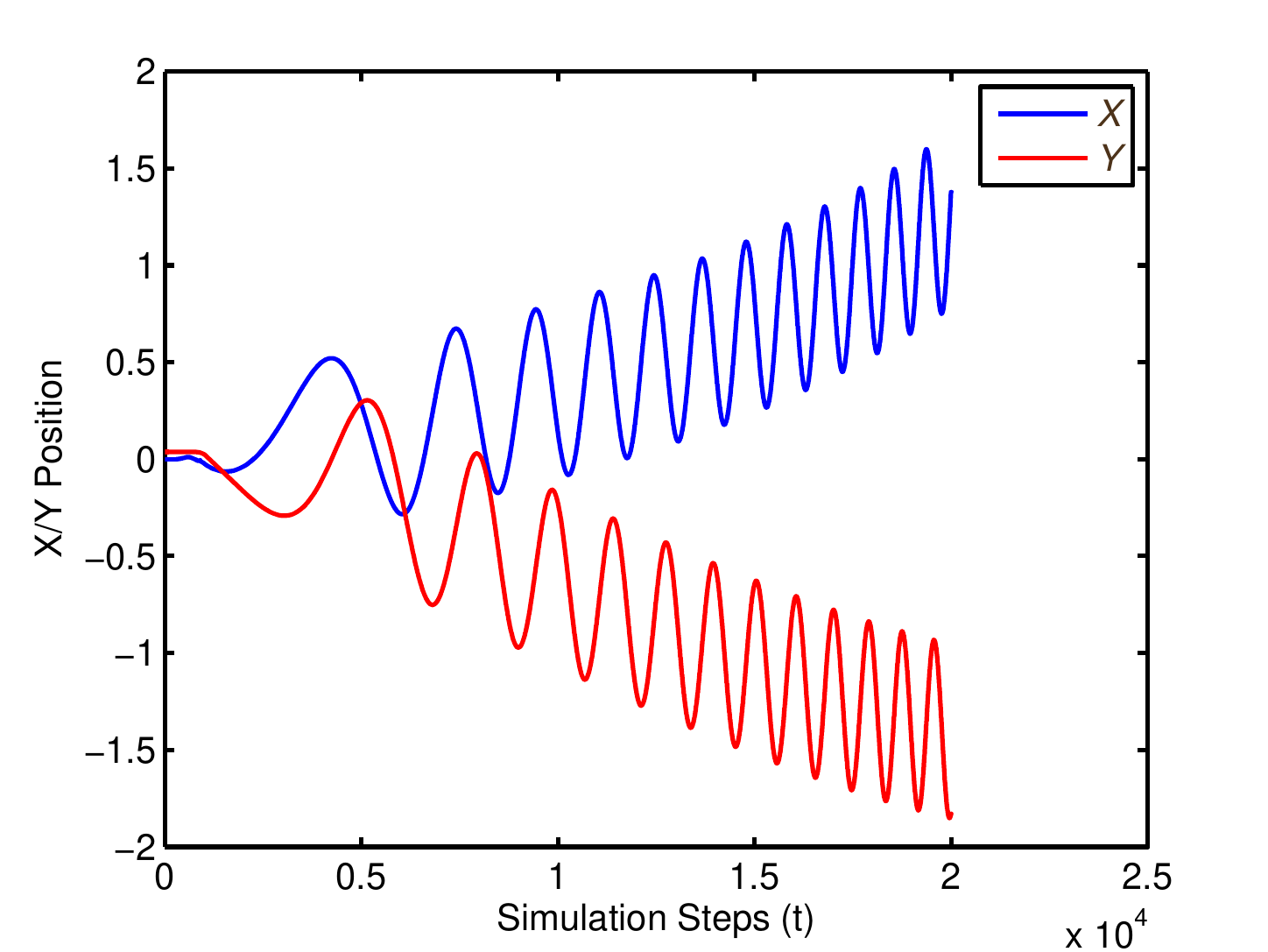}}
\subfigure[]{\includegraphics[width=.235\textwidth]{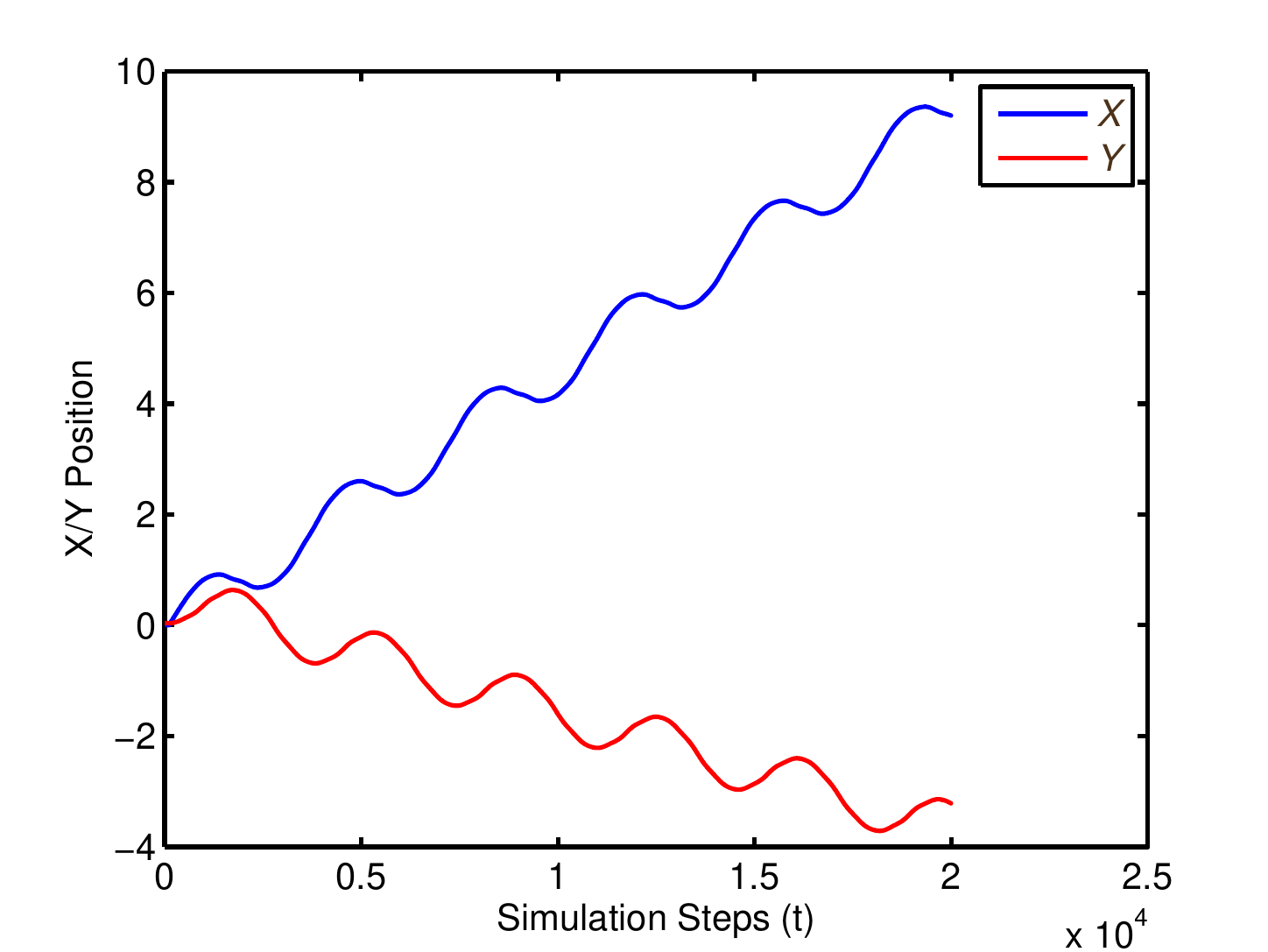}}
\caption{\it (a,b,c,d)- Observation of three links (head, middle and tail) positions for snake build of 13 links. (e,f,g,h) - X,Y dependent on simulation time. \textbf{(a),(e)} Phase shift of $\pi/2$ applied to horizontal gait. \textbf{(b),(f)} Phase shift of $\pi/2$ applied to vertical gate. \textbf{(c),(g)} Phase shift of $\pi/3$ applied to horizontal gait. \textbf{(d),(h)} Phase shift of $\pi/3$ applied to vertical gait.}
\label{fig:TwistGate}
\end{figure}

In Fig.~\ref{fig:TwistGate} we plot positions and phase shifts of different segments of a snake's body. It is well visible that boundary and middle segments have different dynamics related to the amplitude and phase of the movement. The rhythmic drivers shown in Figs.~\ref{fig:SimulinkModel} and \ref{Fig:CPG} have typically homogeneous structure and well correspond to the CPG/CML structure shown by Eq.(\ref{eq:rhythmic}), however the dynamics has an inhomogeneous form. Considering the snake robot from Fig.~\ref{fig:Hardware}, we can attribute these differences in dynamics to physical constraints. Despite these constraints are the same for all segments, e.g. weakness of servomotors, they affect differently boundary and internal segments. Additionally, individual differences of modules, such as unsymmetrical distribution of masses or mechanical differences of modules, create some small variations of dynamics.

Returning now to the analytical modeling of a rhythmic dynamics in the way of Eq.(\ref{eq:rhythmic}), we need to agree whether impact of constraints can be associated with the periodic generator $\dot {\underline x} = \underline f (\underline x, \underline \alpha)$ or with the coupling term $\phi(\underline x, \underline \beta)$. Since the periodic generator is independent of the spatial position of the segment, only the coupling term can be in charge of constraints. This interpretation makes also sense because all generators without coupling demonstrate the same unperturbed dynamics, and only coupling introduces perturbations. Thus, we need to agree that couplings of Eq.(\ref{eq:rhythmic}) should be heterogeneous to reflect an implicit impact of physical constraints.   However, the approach well-known in the CML community, e.g.~\cite{Atmanspachera05} or \cite{Jost02}, introduces primarily homogeneous couplings, this is a basis for several essential features of CML, e.g. a high scalability of coupled oscillators or multiple synchronization effects appearing in them. Obviously, that further investigations should be undertaken to express heterogeneous character of rhythmic control in multi-body systems.

\section{CONCLUSION}
\label{sec:Conclusion}

In this paper we introduced an algorithmic framework for adaptive locomotion of snake-like robot organisms. The goal was to investigate a dynamics of such a multi-body system and to develop an algorithm, which can target a goal while avoiding the obstacles in a energy efficient manner. Experiments are performed in simulation and with real $25$ DOFs snake robot. Several more theoretical issues related to homogeneous structure of CPG/CML-like models are considered, where we estimated that a proper model for a rhythmic control of a multi-body system should involve heterogeneous couplings. This is a preliminary conclusion, which should be considered more in detail, since this would essentially impact scalability and synchronization effect of rhythmic gait generators.

\section{ACKNOWLEDGEMENTS}
The ``SYMBRION'' project is funded by the European Commission within the work programme ``Future and Emergent Technologies Proactive'' under the grant agreement no. 216342. The ``REPLICATOR'' project is funded within the work programme ``Cognitive Systems, Interaction, Robotics'' under the grant agreement no. 216240.

\end{document}